\documentclass[letterpaper]{article} 
\usepackage{aaai2026}  
\usepackage{times}  
\usepackage{helvet}  
\usepackage{courier}  
\usepackage[hyphens]{url}  
\usepackage{graphicx} 
\usepackage{natbib}  
\usepackage{caption} 
\usepackage{algorithm}
\usepackage{algorithmic}

\usepackage{newfloat}
\usepackage{listings}
\DeclareCaptionStyle{ruled}{labelfont=normalfont,labelsep=colon,strut=off} 
\floatstyle{ruled}
\newfloat{listing}{tb}{lst}{}
\floatname{listing}{Listing}
\usepackage{diagbox}
\usepackage{booktabs}
\usepackage{upgreek} 
\usepackage{bm}
\usepackage{amsmath} 
\usepackage{amsfonts}
\usepackage{multirow}
\usepackage{makecell}
\usepackage{colortbl} 
\usepackage{arydshln}
\definecolor{colorFst}{rgb}{0.752,0.89,0.792}          
\definecolor{colorSnd}{rgb}{0.886,0.929,0.725}      
\definecolor{colorTrd}{rgb}{1,0.980,0.757}     

\title{SmartSplat: Feature-Smart Gaussians for Scalable Compression of Ultra-High-Resolution Images}
\author {
   Linfei Li\textsuperscript{\rm 1},
    Lin Zhang\textsuperscript{\rm 1}\thanks{Corresponding Author},
    Zhong Wang\textsuperscript{\rm 2},
    Ying Shen\textsuperscript{\rm 1}
}
\affiliations {
    \textsuperscript{\rm 1}School of Computer Science and Technology, Tongji University\\
    \textsuperscript{\rm 2} Department of Automation, Shanghai Jiao Tong University\\
    cslinfeili@tongji.edu.cn, cslinzhang@tongji.edu.cn, cszhongwang@sjtu.edu.cn, yingshen@tongji.edu.cn
}

\usepackage{bibentry}

\begin{document}

\maketitle

\begin{abstract}
Recent advances in generative AI have accelerated the production of ultra-high-resolution visual content. However, traditional image formats face significant limitations in efficient compression and real-time decoding, which restricts their applicability on end-user devices. Inspired by 3D Gaussian Splatting, 2D Gaussian image models have achieved notable progress in enhancing image representation efficiency and quality. Nevertheless, existing methods struggle to balance compression ratios and reconstruction fidelity in ultra-high-resolution scenarios. To address these challenges, we propose SmartSplat, a highly adaptive and feature-aware GS-based image compression framework that effectively supports arbitrary image resolutions and compression ratios. By leveraging image-aware features such as gradients and color variances, SmartSplat introduces a Gradient-Color Guided Variational Sampling strategy alongside an Exclusion-based Uniform Sampling scheme, significantly improving the non-overlapping coverage of Gaussian primitives in pixel space. Additionally, a Scale-Adaptive Gaussian Color Sampling method is proposed to enhance the initialization of Gaussian color attributes across scales. Through joint optimization of spatial layout, scale, and color initialization, SmartSplat can efficiently capture both local structures and global textures of images using a limited number of Gaussians, achieving superior reconstruction quality under high compression ratios. Extensive experiments on DIV8K and a newly created 16K dataset demonstrate that SmartSplat significantly outperforms state-of-the-art methods at comparable compression ratios and surpasses their compression limits, exhibiting strong scalability and practical applicability. This framework can effectively alleviate the storage and transmission burdens of ultra-high-resolution images, providing a robust foundation for future high-efficiency visual content processing. The code is publicly available at https://github.com/lif314/SmartSplat.
\end{abstract}

\section{Introduction}
With the rapid development of generative artificial intelligence, Ultra-High-Resolution (UHR) visual content has become increasingly accessible and widely disseminated  \cite{zhang2025diffusion4k, ren2024ultrapixel}. However, the resulting high-resolution image data poses significant challenges for storage and transmission, necessitating image representations that offer both high compression ratios and efficient decoding. Traditional formats such as PNG \cite{png2006} and JPEG \cite{jpeg1991} exhibit notable limitations in this context; for instance, JPEG typically achieves a maximum compression ratio of around 50×, which falls short of meeting the demands for efficient transmission and real-time rendering of ultra-high-resolution imagery.

Implicit Neural Representations (INRs) have attracted substantial research interest due to their powerful compression capabilities enabled by neural networks. Nevertheless, existing INR-based methods \cite{siren2020, Beyond2022, fourier2020} generally rely on fixed architectures and full-image training to preserve visual fidelity. These methods require intensive computational resources for ultra-high-resolution images, limiting scalability. Furthermore, the dependency on neural inference leads to slow decoding, making such methods less suitable for real-time applications that demand both rapid decoding and dynamic quality adjustment.

\begin{figure}
    \centering
    \includegraphics[width=0.5\textwidth]{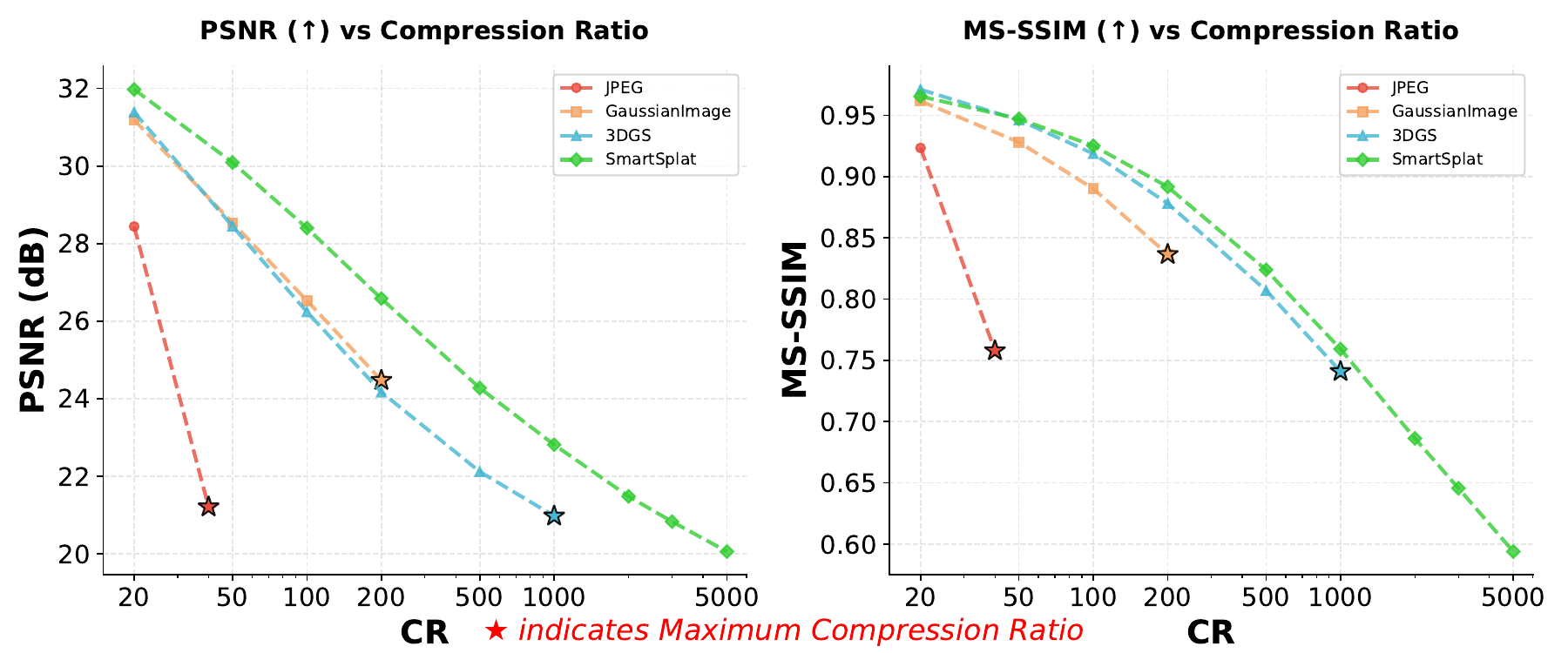}
    \caption{Comparison with baselines on $4218\times 7350$ image. SmartSplat consistently outperforms baselines under the same Compression Ratio (CR) and surpasses the maximum compression limits achieved by previous approaches, maintaining high-fidelity reconstruction even at extreme compression levels (e.g., 1000×).
}
    \label{fig:title_teaser}
    \vspace{-10px}
\end{figure}

\begin{figure*}
    \centering
    \includegraphics[width=0.99\textwidth]{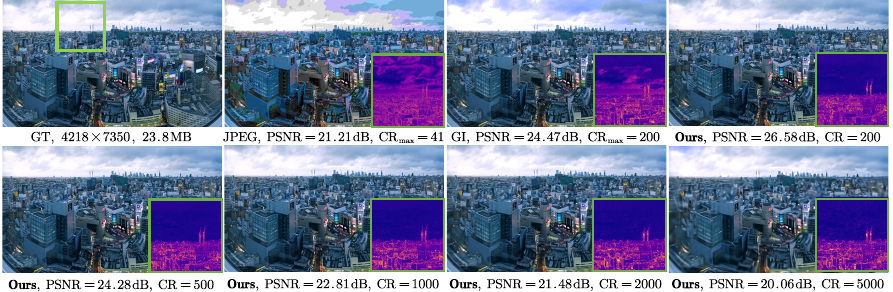}
    \caption{SmartSplat maintains visual quality under extreme high compression ratios. Under maximum compression ratio ($\mathrm{CR}_{\text{max}}$), JPEG shows severe artifacts, and GI struggles with scalability. In contrast, SmartSplat outperforms GI at the same compression ratio and rivals JPEG even at 2000$\times$, maintaining visually pleasing results up to 5000$\times$. The insets visualize the corresponding error images, with brighter colors indicating higher errors.}
    \label{fig:teaser_big}
    \vspace{-10px}
\end{figure*}

Concurrently, 3D Gaussian Splatting (3DGS) \cite{kerbl3Dgaussians} has recently emerged as a novel scene representation technique. By explicitly modeling 3D Gaussian primitives and incorporating a differentiable tile-based rasterization pipeline, it achieves a compelling balance between rendering quality and real-time performance. Inspired by this paradigm, several studies \cite{zhang2024gaussianimage, zhang2024imagegs, zhu2025lig} have extended 3DGS to 2D image representation, proposing spatially-aware 2D Gaussian modeling and rendering frameworks that substantially improve training and decoding efficiency. However, these methods typically rely on a large number of Gaussian primitives to ensure reconstruction accuracy or only achieve limited compression ratios on low-resolution images (typically below 2K), thus falling short of the efficiency requirements for ultra-high-resolution image compression in practical applications.

Accordingly, our research aims to develop a high-compression-ratio representation framework tailored for ultra-high-resolution images at 8K, 16K, and beyond. Such images typically reach sizes ranging from tens to hundreds of megabytes, posing significant challenges for storage, transmission, and sharing. Therefore, there is an urgent need for a compact and efficient representation method that balances compression efficiency with reconstruction quality.

To fill this gap, we propose SmartSplat, a feature-driven 2D Gaussian image compression framework. We begin by analyzing the relationship between compression ratios and the density of Gaussian representations, highlighting that high compression ratios inherently constrain the number of Gaussian points, thereby increasing the difficulty of faithful image reconstruction.

To mitigate this challenge, SmartSplat introduces a highly adaptive Gaussian distribution strategy guided by image features. Specifically, it introduces Gradient-Color Guided Variational Sampling and Exclusion-based Uniform Sampling to jointly optimize the means and scales of Gaussians, while a Scale-Adaptive Color Initialization scheme is proposed to enhance the expressiveness of limited Gaussian primitives in capturing both local structures and global textures. This design enables high-quality reconstruction under strict compression budgets, making it well-suited for practical applications in high-resolution scenarios.

Furthermore, to evaluate the performance of SmartSplat on ultra-high-resolution images, we construct a 16K image dataset, termed DIV16K, by leveraging the Aiarty Image Enhancer tool. As shown in Figures \ref{fig:title_teaser} and \ref{fig:teaser_big}, extensive experiments conducted on both 8K and 16K images reveal that SmartSplat not only outperforms state-of-the-art methods under the same compression ratios but also surpasses their compression limits, achieving competitive reconstruction quality at significantly higher compression levels.

In summary, our main contributions are as follows:
\begin{itemize}
    \item A unified analysis between UHR image compression ratios and GS-based representation, emphasizing the principal challenges involved.
    \item Development of an adaptive Gaussian sampling strategy that jointly optimizes means, scales, and colors to enable compact and efficient UHR image representations.
    \item Extensive experiments on DIV8K and our newly constructed DIV16K dataset demonstrate that SmartSplat achieves superior image representation quality under high compression ratios, significantly outperforming existing GS-based methods.
\end{itemize}

\section{Related Work}
\subsubsection{Implicit Neural Representation.}
In recent years, Implicit Neural Representations (INRs) have demonstrated significant potential in the domains of image modeling and compression. Early approaches \cite{siren2020, fourier2020, Beyond2022, fathony2021multiplicative, wire2023, Li_Zhang_Wang_Zhang_Li_Shen_2025} typically employ multilayer perceptrons (MLPs) to directly regress pixel values, leveraging positional encoding to enhance representational capacity. However, these methods often suffer from slow training and high inference costs, particularly when dealing with high-resolution images. To address these limitations, subsequent studies introduce spatial priors through structured feature grids, such as hierarchical grids \cite{chen2023neurbf, Acorn2021, takikawa2021nglod} and hash-based encodings \cite{ngp2022}, which alleviate the burden on MLPs and substantially accelerate training while maintaining reconstruction quality. Nevertheless, these techniques remain memory-intensive and struggle to adapt to fine-grained and spatially varying image details.

\subsubsection{GS-based Image Representation.}
3D Gaussian Splatting (3DGS) \cite{kerbl3Dgaussians} has emerged as a promising paradigm for view synthesis \cite{Yu2024MipSplatting, Huang2DGS2024, lee2024c3dgs} and reconstruction \cite{li2024gs3lam, MatsukimonogsCVPR2024}, offering exceptional controllability and real-time rendering capabilities through its differentiable tile-based rasterization mechanism and explicit 3D Gaussian representations. Building on this foundation, GaussianImage \cite{zhang2024gaussianimage} extends the principles of 3DGS to the 2D image domain by adapting Gaussian primitives to planar image space for image fitting and compression. While the method achieves satisfactory visual quality, its reliance on a two-stage optimization pipeline and computationally expensive vector quantization \cite{Bhalgat2020vq} introduces significant efficiency bottlenecks. Further advancing this direction, the LIG \cite{zhu2025lig} framework employs a hierarchical Gaussian fitting strategy for high-resolution image reconstruction. However, it prioritizes fitting accuracy over compression performance and requires a large number of Gaussian components. Furthermore, ImageGS \cite{zhang2024imagegs} introduces a content-aware initialization scheme along with a progressive training strategy to enhance optimization efficiency. Nevertheless, it remains suboptimal in extreme-rate image compression scenarios, particularly for ultra-high-resolution images.

\begin{figure*}
    \centering
    \includegraphics[width=0.98\textwidth]{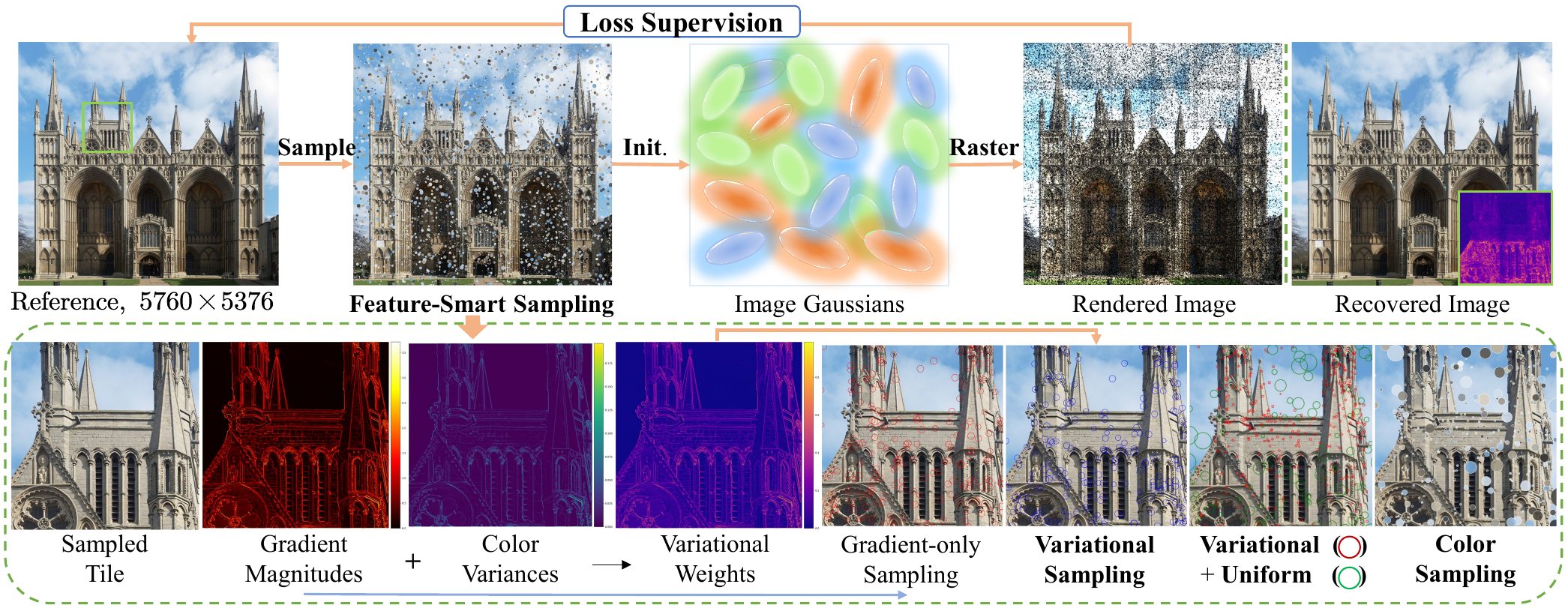}
    \caption{Pipeline of SmartSplat. Given an input image, SmartSplat initializes Gaussian primitives via feature-aware sampling and optimizes them through differentiable rasterization to learn compact, perceptually-aware representations.}
    \label{fig:teaser_pipeline}
    \vspace{-10px}
\end{figure*}

\section{Methodology}
\subsection{Preliminaries: Gaussian Image Splatting}
3DGS \cite{kerbl3Dgaussians} represents a 3D scene using a set of anisotropic Gaussian distributions in 3D space. Each Gaussian is parameterized by its mean position, scale, rotation, opacity, and color. During rendering, these Gaussians are projected onto the image plane through a tile-based rasterization pipeline, resulting in 2D elliptical splats. Then, a front-to-back $\alpha$-blending operation is applied at the pixel level to synthesize novel views.

Extending the 3DGS paradigm to the 2D domain allows for image representation using 2D Gaussian primitives. Specifically, each 2D Gaussian is defined by a mean vector $\boldsymbol{\upmu} \in \mathbb{R}^{2}$, a 2D covariance matrix $\mathbf{\Sigma} \in \mathbb{R}^{2\times2}$, a color vector $\mathbf{c} \in \mathbb{R}^{3}$, and an opacity value $o \in \mathbb{R}$. The contribution of a Gaussian at a given pixel location $\mathbf{x}$ is given by:
\begin{align}
G(\mathbf{x}) = \exp\left(-\frac{1}{2}(\mathbf{x}-\boldsymbol{\upmu})^{T}\mathbf{\Sigma}^{-1}(\mathbf{x}-\boldsymbol{\upmu})\right),
\label{eq:2d-gaussian}
\end{align} 
where the covariance matrix $\mathbf{\Sigma}$ must be positive semi-definite. Direct optimization of $\mathbf{\Sigma}$ via gradient descent, as in LIG \cite{zhu2025lig}, does not guarantee this property. GaussianImage \cite{zhang2024gaussianimage} adopts a Cholesky decomposition \cite{Cholesky2009} approach, where the covariance matrix $\mathbf{\Sigma}$ is factorized as the product of a lower triangular matrix $\mathbf{L} \in \mathbb{R}^{2\times2}$ and its transpose:
\begin{equation}
\mathbf{\Sigma} = \mathbf{L} \mathbf{L}^T.
\label{eq:cholesky}
\end{equation}
Alternatively, inspired by 3DGS, the covariance matrix can also be expressed as the product of a rotation matrix $\mathbf{R} \in \mathbb{R}^{2\times2}$ and a scaling matrix $\mathbf{S} \in \mathbb{R}^{2\times2}$:
\begin{align}
\mathbf{\Sigma} = \mathbf{R} \mathbf{S} \mathbf{S}^{T}  \mathbf{R}^{T}.
\label{eq:2d-gaussian-covariance}
\end{align}

Furthermore, since rendering on the image plane does not require depth sorting of Gaussian primitives, the color at each pixel $\hat{\mathbf{c}}(\mathbf{x})$ can be computed via a forward-style $\alpha$-blending over the $\mathcal{N}$ overlapping Gaussians:
\begin{equation}
    \hat{\mathbf{c}}(\mathbf{x}) =\sum_{i \in \mathcal{N}} \mathbf{c}_i \cdot o_i \cdot G_i(\mathbf{x}) \prod_{j=1}^{i-1} \left(1 - o_j G_j(\mathbf{x})\right),
\end{equation} where $o_i$ denotes the opacity, $\mathbf{c}_i$ is the color coefficient, and $G_i(\mathbf{x})$ represents the value of the $i$-th 2D Gaussian evaluated at location $\mathbf{x}$. This formulation accumulates contributions from all overlapping Gaussians in a fixed order without requiring explicit visibility reasoning.

\subsection{Feature-Smart Gaussians}
\subsubsection{Problem Formulation.} 
Assume an original $H \times W$ RGB image is encoded with 8 bits per channel (excluding transparency), resulting in an uncompressed size of $3HW$ bytes. Since pixel rendering is independent of the ordering of Gaussians, we assume a constant opacity of 1. Under this assumption, each Gaussian primitive requires seven 32-bit floating-point parameters—representing position, scale, rotation, and color—amounting to approximately 32 bytes per primitive. With vector quantization-based compression techniques \cite{Bhalgat2020vq, zhang2024gaussianimage}, the storage per primitive can be reduced to 7 bytes. Given a target compression ratio $\mathrm{CR}$, the maximum number of Gaussian primitives $N_g$ allowed is determined by:
\begin{equation}
    N_g = \frac{3HW}{7 \cdot \mathrm{CR}}.
    \label{eq:cr_gs}
\end{equation}

\subsubsection{Gaussian Image Representation Decomposition.}
To gain a deeper understanding of the relationship between Gaussian distributions and the image space, we adopt a physically interpretable covariance decomposition as defined in Eq. \ref{eq:2d-gaussian-covariance}. Accordingly, for each Gaussian element representing the image, we normalize its color component $\mathbf{c}$ to the range $[0, 1]$, and fix its opacity $o$ to 1. The associated rotation matrix $\mathbf{R}$ can be parameterized as:
\begin{equation}
    \mathbf{R} = \begin{bmatrix}
\cos(\theta) & -\sin(\theta) \\
\sin(\theta) & \cos(\theta)
\end{bmatrix},
\end{equation} where $\theta \in [0, 2\pi)$ denotes the rotation angle. During initialization, $\theta$ is sampled from the interval $[0,1)$ and scaled by $2\pi$ to span the full angular range. The scale matrix of the Gaussian is defined as a symmetric matrix:
\begin{equation}
    \mathbf{S} = \begin{bmatrix}
        s_x & 0 \\
        0 & s_y
\end{bmatrix},
\end{equation} where $s_x$ and $s_y$ denote the scales along the $x$- and $y$-axes, respectively. Following the $3\sigma$ rule, the maximum influence radius $R_{g \rightarrow p}$ of a Gaussian on neighboring pixels during rasterization can be approximated (ignoring rotation) by:
\begin{equation}
    R_{g \rightarrow p} \approx 3 \cdot \max(s_x, s_y).
    \label{eq:radius_to_scale}
\end{equation}

Based on the aforementioned decomposition analysis and guided by Eq. \ref{eq:cr_gs}, it can be seen that the central challenge of representing an image using limited Gaussians under a given compression ratio lies in the effective design of their spatial distribution. This desired design must adaptively capture both high- and low-frequency structures within the image. To achieve this, we propose a feature-guided joint sampling strategy that simultaneously considers the position, scale, and color attributes of Gaussians, enabling efficient representation of images at arbitrary resolutions and compression ratios. Specifically, as shown in Fig. \ref{fig:teaser_pipeline}, image features such as gradients and color variations are leveraged through a Gradient-Color Guided Variational Sampling strategy and an Exclusion-based Uniform Sampling scheme, which collectively guide the initialization of the means and scales of Gaussians to ensure non-overlapping and content-aware spatial coverage. In addition, a Scale-Adaptive Gaussian Color Sampling strategy is introduced to initialize the color attributes of Gaussians across scales, further enhancing the fidelity of representation. This unified design allows SmartSplat to capture both fine-grained local structures and coarse global patterns using a compact set of feature-aware Gaussians, thereby achieving high-quality image reconstruction under high compression ratios.

\subsubsection{Gradient-Color Guided Variational Sampling.}
Intuitively, high-frequency regions in an image should be represented using densely distributed Gaussians with smaller scales, while low-frequency regions are more appropriately modeled with sparsely distributed Gaussians of larger scales. A straightforward approach is to select Gaussian positions based solely on image gradients \cite{zhang2024imagegs}. However, since gradients primarily emphasize structural information, this may result in inadequate representation of low-frequency regions when the number of Gaussian primitives is limited. To address this issue, we propose a variational sampling strategy that jointly leverages both image gradients and color variance. Gradients are employed to guide denser sampling in structure-rich areas, while color variance is used to identify regions with high chromatic complexity. Such a joint strategy enables adaptive sampling across different frequency components of the image.

To efficiently process high-resolution images while ensuring uniform coverage during Gaussian initialization, we propose an adaptive step-size block-wise variational sampling strategy. This approach partitions the large-scale image into multiple overlapping or adjacent tiles, within which variational sampling is conducted independently. The adaptive step-size mechanism further guarantees uniform coverage across the entire image.

Specifically, within each tile sub-image $\mathbf{I}_{i,j}$, the local gradient magnitude and color variance of its pixels are computed as follows:
\begin{equation}
    \begin{aligned}
m_{i,j}(\mathbf{x}) &= \frac{1}{C} \sum_{c=1}^C \left\| \nabla \mathbf{I}_{i,j,c}(\mathbf{x}) \right\|_2, \\
v_{i,j}(\mathbf{x}) &= \frac{1}{C} \sum_{c=1}^C \mathrm{Var}\bigl(\mathbf{I}_{i,j,c}(\mathcal{N}_{\mathbf{x}})\bigr),
\end{aligned}
\end{equation}
where $C$ denotes the number of channels, and $\mathcal{N}_{\mathbf{x}}$ represents the neighborhood of pixel $\mathbf{x}$. To eliminate scale discrepancies, the gradient magnitude and color variance are normalized within the tile, yielding $\tilde{m}_{i,j}(\mathbf{x})$ and $\tilde{v}_{i,j}(\mathbf{x})$, respectively. The sampling weight is then defined as a weighted combination of these normalized values:
\begin{equation}
    w_{i,j}(\mathbf{x}) = \lambda_m \cdot \tilde{m}_{i,j}(\mathbf{x}) + (1- \lambda_m) \cdot \tilde{v}_{i,j}(\mathbf{x}),
    \label{eq:vs_weight}
\end{equation}
where $\lambda_m$ denotes the weighting coefficient that balances the contributions of gradient magnitude and color variance.

Based on these sampling weights, the sampling probability of pixel $\mathbf{x}$ within tile $(i,j)$ is given by:
\begin{equation}
    \mathbb{P}_{i,j}(\mathbf{x}) = \frac{w_{i,j}(\mathbf{x})}{\sum_{\mathbf{y} \in \mathbf{I}_{i,j}} w_{i,j}(\mathbf{y})}.
\end{equation}
Finally, multinomial sampling is performed according to this probability distribution to select $n_{i,j}$ points within the tile:
\begin{equation}
    \{\mathbf{x}_k^{(i,j)}\}_{k=1}^{n_{i,j}} \sim \mathrm{Multinomial}\left(n_{i,j}, \{\mathbb{P}_{i,j}(\mathbf{x})\}_{\mathbf{x} \in \mathbf{I}_{i,j}}\right).
\end{equation}

The proposed variational sampling strategy effectively increases the density of samples in regions exhibiting prominent gradients or significant color variation, thereby facilitating a more appropriate initialization of Gaussian primitives.

Evidently, points with higher sampling weights should be assigned smaller scales, while those with lower weights can be allocated larger scales. To ensure spatial smoothness, we adopt an exponential decay function to adaptively compute the scale. Assuming the initial scales along the $x$- and $y$-axes are equal, the scale is given by:
\begin{equation}
    s_{i,j}(\mathbf{x}) = s_{{base}} \cdot \exp(-\frac{1}{2}w_{i,j}(\mathbf{x})).
\end{equation}
To maximize the coverage of the pixel space by Gaussians, we consider the influence radius of a Gaussian (Eq. \ref{eq:radius_to_scale}). Assuming uniform coverage using circles with radius $R_{g \rightarrow p}$, the maximum non-overlapping Gaussian scale can be derived as:
\begin{equation}
    s_{{base}} = \frac{1}{3} R_{g\rightarrow p}= \frac{1}{3} \sqrt{\frac{H W}{ \pi N_g}},
    \label{eq:base_scale}
\end{equation}
where $H, W$ and $N_g$ are defined in Eq. \ref{eq:cr_gs}. This scale initialization strategy adaptively represents images of arbitrary resolutions without relying on any hyperparameters or heuristic clamping.

\subsubsection{Exclusion-based Uniform Sampling.}
Following the variational sampling, an exclusion-based uniform sampling strategy is proposed to ensure adequate coverage of image regions characterized by low structural complexity or minimal color variation.

Specifically, let the set of variationally sampled points be denoted by $\mathcal{X}_{vs}=\{ \mathbf{x}_i^{vs} \}_{i=1}^{N_g^{vs}}$, where $N_g^{vs}$ represents the number of points obtained through variational sampling. During the subsequent uniform sampling stage, the sampled point set $\mathcal{X}_{us}=\{ \mathbf{x}_j^{us} \}_{j=1}^{N_g^{us}}$ must satisfy the following exclusion constraint to ensure spatial separation from the previously selected points:
\begin{equation}
    \forall j, \quad \min_{i} \left\| \mathbf{x}_j^{us} - \mathbf{x}_i^{vs} \right\| \geq r_{{excl}},
\end{equation}
where $r_{{excl}}$ denotes the exclusion radius. To prevent excessive overlap between variationally sampled and uniformly sampled points, the exclusion radius is determined by incorporating both the Gaussian influence radius and the scale of variationally sampled points. Concretely, $r_{{excl}}$ is defined as the maximum of the base scale and the median scale of the variational samples:
\begin{equation}
    r_{{excl}} = \max\left(s_{{base}}, \text{median}(\{s_i^{vs}\}_{i=1}^{N_g^{vs}})\right),
\end{equation} 
where $s_{{base}}$ is defined as in Eq. \ref{eq:base_scale}.

Furthermore, to ensure that Gaussian kernels adequately cover the entire pixel domain of the image, we adopt a Query-to-Reference KNN algorithm to estimate the scale of uniformly sampled points. Specifically, let the complete set of points be denoted by $\mathcal{X} = \mathcal{X}_{{vs}} \cup \mathcal{X}_{{us}}$, where $\mathcal{X}_{{vs}}$ and $\mathcal{X}_{{us}}$ represent the variationally sampled and uniformly sampled point sets, respectively. For each uniform sample $\mathbf{x}_j^{{us}} \in \mathcal{X}_{{us}}$, a $K$-nearest neighbor search is performed within the complete set $\mathcal{X}$ to determine its local scale. The scale is defined as:
\begin{equation}
    s_j^{{us}} = \sqrt{ \frac{1}{K} \sum_{\mathbf{q} \in \mathcal{N}_K(\mathbf{x}_j^{{us}}, \mathcal{X})} \left\| \mathbf{x}_j^{{us}} - \mathbf{q} \right\|^2 },
    \label{eq:knn_scale}
\end{equation}
where $\mathcal{N}_K(\mathbf{x}_j^{{us}}, \mathcal{X})$ denotes the set of $K$ nearest neighbors of $\mathbf{x}_j^{{us}}$ within $\mathcal{X}$. The resulting scale $s_j^{{us}}$ reflects the local sampling density around the point. This approach adaptively enhances coverage in sparse regions, thereby improving the overall robustness and representational fidelity of the sampling distribution across the image domain.

\subsubsection{Scale-Adaptive Gaussian Color Sampling.}
Following the initialization of 2D Gaussian positions and scales via variational and uniform sampling, we introduce a scale-adaptive Gaussian-weighted median sampling strategy to estimate the color parameters of each Gaussian element. This approach aims to enhance structural fidelity and improve robustness to local noise and outliers. Unlike traditional methods based on random initialization \cite{zhang2024gaussianimage, zhu2025lig} or pixel-center estimation \cite{zhang2024imagegs}, the proposed strategy effectively combines the robustness of median estimation with the spatial sensitivity of Gaussian weighting. This enables more accurate color recovery in regions with high-frequency textures or abrupt depth changes, thereby improving reconstruction quality and perceptual consistency while accelerating convergence.

Specifically, for each sampled point $\mathbf{x}_i \in \mathcal{X} = \mathcal{X}_{{vs}} \cup \mathcal{X}_{{us}}$, obtained from either variational sampling or uniform sampling, and associated with a scale parameter $s_i$, we define a circular neighborhood $\mathcal{N}_{\mathbf{x}_i}$ centered at $\mathbf{x}_i$ with radius $s_i$:
\begin{equation}
    \mathcal{N}_{\mathbf{x}_i} = \left\{ \mathbf{u} \in \mathbb{Z}^2 \,\middle|\, \|\mathbf{u} - \mathbf{x}_i\|_2 \leq s_i \right\}.
\end{equation}
For each pixel $\mathbf{u} \in \mathcal{N}_{\mathbf{x}_i}$, a spatial weight is assigned based on a 2D isotropic Gaussian kernel centered at $\mathbf{x}_i$:
\begin{equation}
    w_i(\mathbf{u}) = \exp\left(-\frac{\|\mathbf{u} - \mathbf{x}_i\|^2}{2 \sigma_i^2} \right), \quad \text{where} \;\; \sigma_i = s_i.
\end{equation}
Then, the RGB color $\mathbf{c}_i \in \mathbb{R}^3$ corresponding to point $\mathbf{x}_i$ is estimated via a Gaussian-weighted median over the pixel intensities $\mathbf{I}(\mathbf{u})$ within its neighborhood. Specifically, for each color channel $d \in \{1, 2, 3\}$, the channel value is determined by solving the following minimization:
\begin{equation}
    \mathbf{c}_i^{(d)} = \arg\min_{z \in \mathbb{R}} \sum_{\mathbf{u} \in \mathcal{N}_{\mathbf{x}_i}} w_i(\mathbf{u}) \cdot \left| z - \mathbf{I}^{(d)}(\mathbf{u}) \right|.
\end{equation}

This scale-adaptive color sampling strategy leverages the spatial coherence inherent in Gaussian sampling while incorporating the robustness of median filtering to reduce sensitivity to outlier scales. As a result, it enables accurate color estimation across multiple scales, even in regions affected by noise or exhibiting significant local variations.

\begin{table*}[ht]
  \centering
  \resizebox{\textwidth}{!}{
  \begin{tabular}{cc|cccccc}
    \hline
    Dataset & $\mathrm{CR}$ & \textit{3DGS} & \textit{LIG} & \textit{GI (RS)} & \textit{GI (Cholesky)} & \textit{ImageGS} & \textit{SmartSplat (Ours)} \\
    \hline
    \multirow{6}{*}{DIV8K}
    & 20   & 30.99 / 0.9636 & 28.05 / 0.9362 & 30.45 / 0.9707 & 30.33 / 0.9698 & 32.00 / 0.8680 & \textbf{33.26 / 0.9752} \\
    & 50   & 28.56 / 0.9340 & 24.90 / 0.8402 & 26.99 / 0.9291 & 26.87 / 0.9271 & 29.47 / 0.8052 & \textbf{29.65 / 0.9482} \\
    & 100  & 26.84 / 0.8990 & 22.91 / 0.7230 & 25.00 / 0.8827 & 24.90 / 0.8790 & 26.65 / 0.7449 & \textbf{27.49 / 0.9164} \\
    & 200  & 24.92 / 0.8556 & 21.06 / 0.5792 & 23.45 / 0.8223 & 23.35 / 0.8176 & \textbf{26.80} / 0.7181 & 25.75 / \textbf{0.8745} \\
    & 500  & 22.38 / 0.7874 & 17.68 / 0.3633 & Fail & Fail & \textbf{24.88} / 0.6544  & 23.82 /  \textbf{0.8055} \\
    & 1000 & 20.38 / 0.7068 & 12.49 / 0.2083 & Fail & Fail & \textbf{23.50} / 0.6165 & 22.66 / \textbf{0.7469} \\
    \hline
    \multirow{7}{*}{DIV16K}
    & 50   & OOM & 24.42 / 0.4561 & 29.24 / 0.7917 & 29.14 / 0.7899 & OOM & \textbf{34.34 / 0.9267} \\
    & 100  & OOM & 21.37 / 0.3815 & 27.39 / 0.7648 & 27.28 / 0.7623 & OOM & \textbf{33.00 / 0.9117} \\
    & 200  & OOM & 18.01 / 0.3171 & 25.63 / 0.7394 & 25.51 / 0.7365 & OOM & \textbf{31.85 / 0.8897} \\
    & 500  & 28.61 / 0.8117 & 11.97 / 0.2015 & Fail & Fail & OOM & \textbf{29.40 / 0.8524} \\
    & 1000 & 27.06 / 0.7854 & 6.78 / 0.1749 & Fail & Fail & OOM & \textbf{27.49 / 0.8226} \\
    & 2000 & 25.54 / 0.7642 & Fail & Fail & Fail & OOM & \textbf{25.70 / 0.7966} \\
    & 3000 & Fail & Fail & Fail & Fail & Fail & \textbf{24.72 / 0.7844} \\
    \hline
  \end{tabular}
  }
  \caption{Quantitative results on DIV8K (Avg. Res./Size: $5736\times 6120$/53.56\text{MB}) and DIV16K (Avg. Res./Size: $12684\times 15898$/235.52MB). Each cell reports PSNR / MS-SSIM (DIV8K) or PSNR / SSIM (DIV16K). ``OOM'' denotes out-of-memory, and ``Fail'' means training failure due to insufficient Gaussians.}
  \label{tab:combined-compression-results}
\end{table*}

\begin{figure*}
    \centering
    \includegraphics[width=0.99\textwidth]{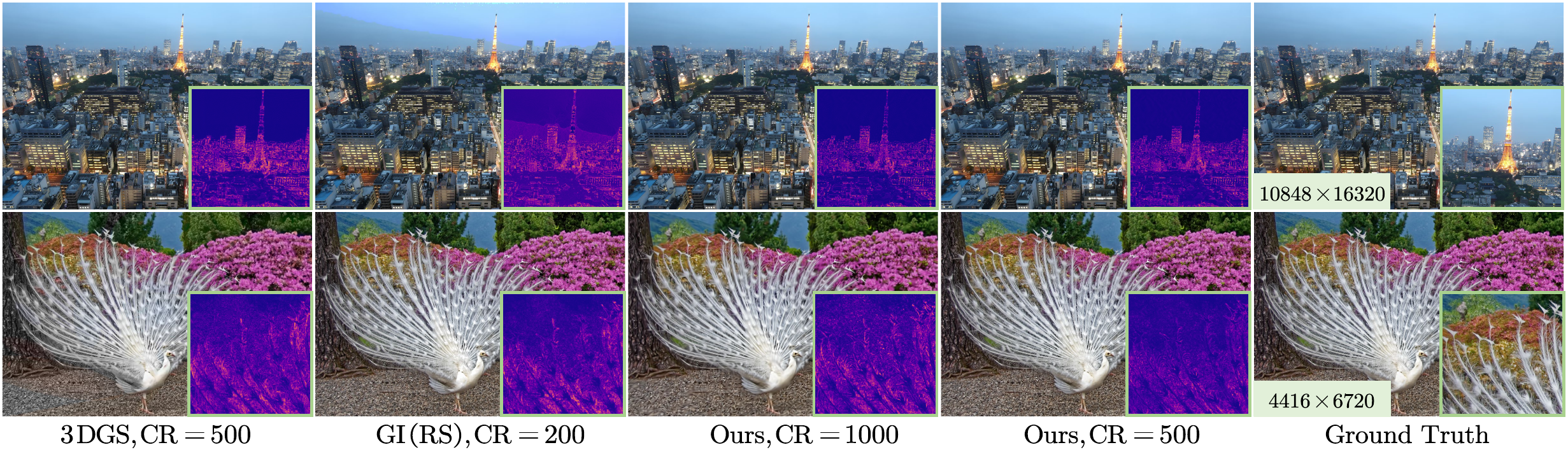}
    \caption{Qualitative comparison on DIV8K and DIV16K.}
    \label{fig:qualitative_comparison}
    \vspace{-10pt}
\end{figure*}

\subsubsection{Optimization.}
Given an input image of size $H \times W$ and a target compression ratio $\mathrm{CR}$, the maximum allowable number of Gaussian elements, denoted by $N_g$, can be computed based on Eq. \ref{eq:cr_gs}. Assuming a variational sampling ratio of $\lambda_g$, the numbers of Gaussians allocated to variational sampling and uniform sampling are determined as:
\begin{equation}
    N_g^{vs} = \lambda_g N_g, \quad N_g^{us} = (1 - \lambda_g) N_g.
    \label{eq:sample_num}
\end{equation}

Using the aforementioned sampling strategy, a corresponding number of sample points, along with their scales and colors, are used to initialize the Gaussians. Then, the reconstruction process is optimized by minimizing a composite loss that combines the $l_1$ distance and the SSIM \cite{Zhou2004ssim} between the rendered image and the ground-truth image. The overall loss function is defined as:
\begin{equation}
    L = \lambda_l \| \hat{\mathbf{I}} - \mathbf{I} \|_1 + (1 - \lambda_l) ( 1 - \mathrm{SSIM}(\hat{\mathbf{I}}, \mathbf{I})),
    \label{eq:loss}
\end{equation}
where $\hat{\mathbf{I}}$ denotes the reconstructed image rendered from the set of Gaussians, $\mathbf{I}$ represents the corresponding ground-truth image, and $\lambda_l \in [0,1]$ is a weighting factor that balances the contributions of the two loss terms.

\section{Experiments}
\subsection{Experimental Setup}
\subsubsection{Dataset.}
To comprehensively assess the performance of GS-based image compression on ultra-high-resolution content, the DIV8K dataset \cite{Gu2019div8k} was employed as the primary benchmark. In addition, a new dataset, DIV16K, was constructed by applying $8{\times}$ upsampling to images from DIV2K \cite{div2k2017} using the Aiarty Image Enhancer, thereby simulating high-resolution imagery representative of generative AI outputs. Given the significant computational demands of such data, a subset of 16 images from DIV8K and 8 images from DIV16K was selected for evaluation. All images were stored in lossless PNG format, providing a reliable testbed for examining the trade-off between compression efficiency and perceptual fidelity.

\subsubsection{Implementation.}
To handle UHR image initialization efficiently, all sampling was performed in a tile-based manner. A CUDA-based query-to-reference KNN pipeline was introduced for exclusion sampling and scale estimation. On 16K images, the initialization stage can be completed within $2 \sim 5$ seconds. Variational and uniform sampling used $\lambda_m{=}0.9$ and $K{=}3$, respectively. During training, $\lambda_g{=}0.7$ and $\lambda_l{=}0.9$ were adopted. All Gaussian parameters were jointly optimized with Adam over 50K steps using learning rates of $1e-4$, $5e-3$, $5e-2$, and $1e-3$.

\subsubsection{Evaluation Metrics.}
PSNR measures pixel-level distortion, while MS-SSIM \cite{Wang2003msssim} evaluates perceptual and structural fidelity. For 16K images, we employed FusedSSIM \cite{Mallick2024fusedssim, Zhou2004ssim} to prevent OOM errors during evaluation.

\subsubsection{Baselines.}
Due to the high memory cost of INR-based methods on UHR images, this study focuses on comparisons with GS-based methods, including 3DGS \cite{kerbl3Dgaussians}, GI \cite{zhang2024gaussianimage}, LIG \cite{zhu2025lig}, and ImageGS \cite{zhang2024imagegs}. Both GI variants (RS and Cholesky) are evaluated.

\subsection{Evaluation}
\subsubsection{Image Compression Performance Evaluation.}
As illustrated in Table \ref{tab:combined-compression-results} and Fig. \ref{fig:qualitative_comparison}, the evaluation results on the DIV8K and DIV16K datasets demonstrate that SmartSplat consistently outperforms existing methods in reconstruction quality under equivalent compression ratios ($\mathrm{CR}$). As the compression rate increases, the sparsity of the Gaussian distribution in existing methods often leads to the emergence of NaN values during rasterization, which disrupts the optimization process. Although ImageGS adopts an error-driven strategy by incrementally adding Gaussians, this approach tends to introduce instability when the number of Gaussians is limited. In contrast, SmartSplat employs a highly adaptive initialization strategy for Gaussian distribution, enabling stable and efficient iterative optimization across various image resolutions, even under extremely high compression ratios.

Specifically, on the DIV8K dataset, SmartSplat achieves improvements of 1.53 dB in PSNR and 0.0201 in MS-SSIM over the runner-up method, 3DGS, at the same compression ratio. It is noteworthy that 3DGS projects pixels into 3D space using an identity matrix, leading to slower training and significantly higher memory usage. Compared to the 2D Gaussian baseline GI (RS), SmartSplat achieves a 2.57dB PSNR gain and maintains similar quality at 500$\times$ compression, whereas GI (RS) requires 200$\times$ for comparable results.

On the DIV16K dataset, the advantages of SmartSplat are even more pronounced. At lower compression ratios (20/100/200), 3DGS encounters out-of-memory (OOM) issues and fails to complete training, whereas SmartSplat maintains stable optimization and achieves an average PSNR gain about 5.64 dB over GI (RS). At higher compression ratios (above 200$\times$), both GI and ImageGS fail to converge, while SmartSplat continues to deliver superior reconstruction quality and remains robust even under extremely aggressive compression ratios.

\subsubsection{Optimization Performance Evaluation.}
\begin{figure}
    \centering
    \includegraphics[width=0.43\textwidth]{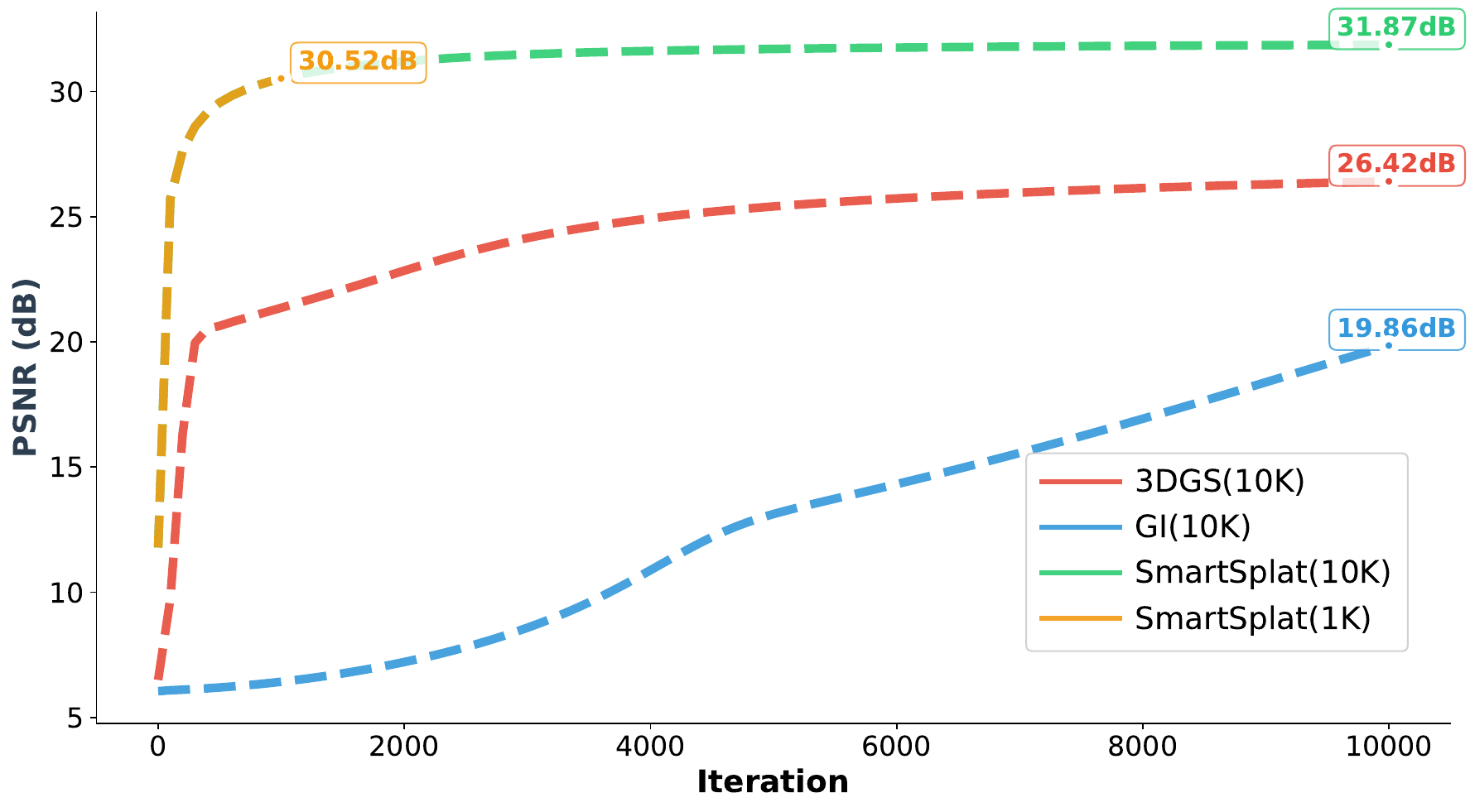}
    \caption{Training convergence speed comparison.}
    \label{fig:convergence}
\end{figure}

The optimization process was evaluated on the $10848 \times 16320$ image shown in Fig. \ref{fig:qualitative_comparison} under a compression ratio of $\mathrm{CR} = 200$. As illustrated in Fig. \ref{fig:convergence}, SmartSplat exhibits a significantly faster convergence rate, attributed to its highly adaptive Gaussian initialization strategy. Notably, it achieves superior reconstruction quality to both 3DGS and GI within only 1K iterations—substantially outperforming their respective results even at 10K iterations. Furthermore, as reported in Table \ref{table:training_speed_16k}, although 3DGS also demonstrates strong image representation capabilities, its memory requirement is approximately $2.56 \times$ that of SmartSplat, and its training time is about $3.49 \times$ longer.  While GI offers certain advantages in training and decoding speed, SmartSplat achieves a 10.66 dB gain in PSNR within just 1K iterations, with training time reduced to 25\% of that of GI, thereby demonstrating a more favorable balance between efficiency and quality.

\begin{table}
  \centering
\resizebox{0.48\textwidth}{!}{
  \begin{tabular}{ccccccc}
    \toprule
     Method & Iter/s $\uparrow$ & TrainTime (s) $\downarrow$ & TrainMem. (GB) $\downarrow$ & FPS $\uparrow$ & PSNR (dB) $\uparrow$ & MS-SSIM $\uparrow$ \\
    \midrule
    3DGS (10K) & 1.32 & 7841.80 & 50.19 & 10.98 & 24.42 & 0.8922  \\
    GI (10K) & 7.44 & 1334.73 & 16.29 & 62.33 & 19.86 & 0.4825 \\
    SmartSplat (10K) & 5.01 & 2237.52 & 19.59 & 32.35 & 31.87 & 0.9354 \\
    SmartSplat (1K) & 5.03 & 336.12 & 19.38 & 33.12 & 30.52 & 0.9209 \\
    \bottomrule
  \end{tabular}
}
  \caption{Training and decoding comparison.}
  \label{table:training_speed_16k}
  \vspace{-10px}
\end{table}
\begin{table}[ht]
  \centering
\resizebox{0.48\textwidth}{!}{
  \begin{tabular}{ccccc}
    \toprule
     Variants & TrainTime (s) $\downarrow$ & PSNR (dB) $\uparrow$ & MS-SSIM $\uparrow$ & FPS $\uparrow$ \\
    \midrule
    Full Random & 456.74 & 22.34 & 0.8435 & 97.39 \\
    +VS/US Means &434.59  & 22.18 & 0.8270 & 90.60  \\
    +VS/US Scales &454.32  & 23.12 & 0.8647 & 92.00  \\
    +SA Colors (Full SmartSplat) & 456.12 & 24.38 & 0.8972  & 94.84 \\
    \bottomrule
  \end{tabular}
}
  \caption{Ablation study. ($\mathrm{CR}=200$, 10K Iterations)}
  \label{table:ablation_study}
  \vspace{-10px}
\end{table}

\subsubsection{Ablation Study.}
To evaluate the impact of SmartSplat’s position, scale, and color initialization strategies, an ablation study was performed on the $4416 \times 6720$ image shown in Fig. \ref{fig:qualitative_comparison}, under a compression ratio of $\mathrm{CR}=200$ and 10K iterations. As shown in Table \ref{table:ablation_study}, the baseline with fully random initialization (Full Random) performs poorly (22.34 dB PSNR, 0.8435 MS-SSIM), revealing its inefficiency in capturing image structure. Adding variational and uniform sampling for mean initialization (+VS/US Means) offers limited PSNR gains but slightly accelerates training. Further integrating scale initialization (+VS/US Scales) significantly boosts performance (23.12 dB PSNR, 0.8647 MS-SSIM), due to better multi-scale adaptation. Finally, incorporating scale-adaptive color initialization (+SA Colors) leads to the full SmartSplat model, achieving 24.38 dB PSNR and 0.8972 MS-SSIM with competitive training time. These results confirm that each component incrementally enhances both reconstruction quality and efficiency.

\section{Conclusion and Future Work}
We proposed SmartSplat, the first GS-based image compression framework that operates effectively on UHR (8K/16K) images. By introducing gradient-color guided variational sampling and exclusion-based uniform sampling, along with scale-adaptive Gaussian color initialization, SmartSplat achieves efficient, non-overlapping Gaussian coverage and strong expressiveness. It outperforms existing methods under the same compression ratios and maintains high reconstruction quality even under extreme high compression ratios.
This study primarily addresses the optimization of Gaussian spatial distribution, with future work targeting advanced attribute compression for improved efficiency.

\section*{Acknowledgments}
This work was supported in part by the National Natural Science Foundation of China under Grants 62272343 and 62476202, and in part by the Fundamental Research Funds for the Central Universities.

\bibliography{aaai2026}

\clearpage
\section{Pipeline of  SmartSplat}
\textit{This section provides a more detailed explanation of the SmartSplat pipeline, serving as a complementary description to Fig. 3 in the main text.}

As illustrated in Fig. \ref{fig:teaser_pipeline}, given an input image, SmartSplat first employs a unified Feature-Smart Sampling strategy to initialize the positions, scales, and color attributes of Gaussian elements, thereby constructing an initial set of 2D Gaussians on the image plane. These Gaussians are then rendered into a synthesized image via a differentiable rasterization process. An optimization objective combining L1 loss and SSIM loss is formulated, and the Gaussian parameters are iteratively refined through gradient descent. After training, the framework yields a high-fidelity reconstructed image representation.

In the Feature-Smart Sampling module, an adaptive step-size block-wise sampling strategy is introduced to efficiently process ultra-high-resolution images while avoiding out-of-memory issues. Within each sampling block, variational sampling is first performed by combining image gradient information and color variance, allowing Gaussians to be preferentially placed in regions with complex structures or significant color variation. To ensure uniform spatial coverage in low-texture regions, an exclusion-based uniform sampling strategy is further employed to balance the distribution of Gaussians. Finally, for each sampled Gaussian, the color attribute is initialized using the Gaussian-weighted median color within its corresponding region. This sampling process is highly adaptive, enabling flexible initialization across arbitrary image resolutions and compression ratios.

\begin{figure*}
    \centering
    \includegraphics[width=0.96\textwidth]{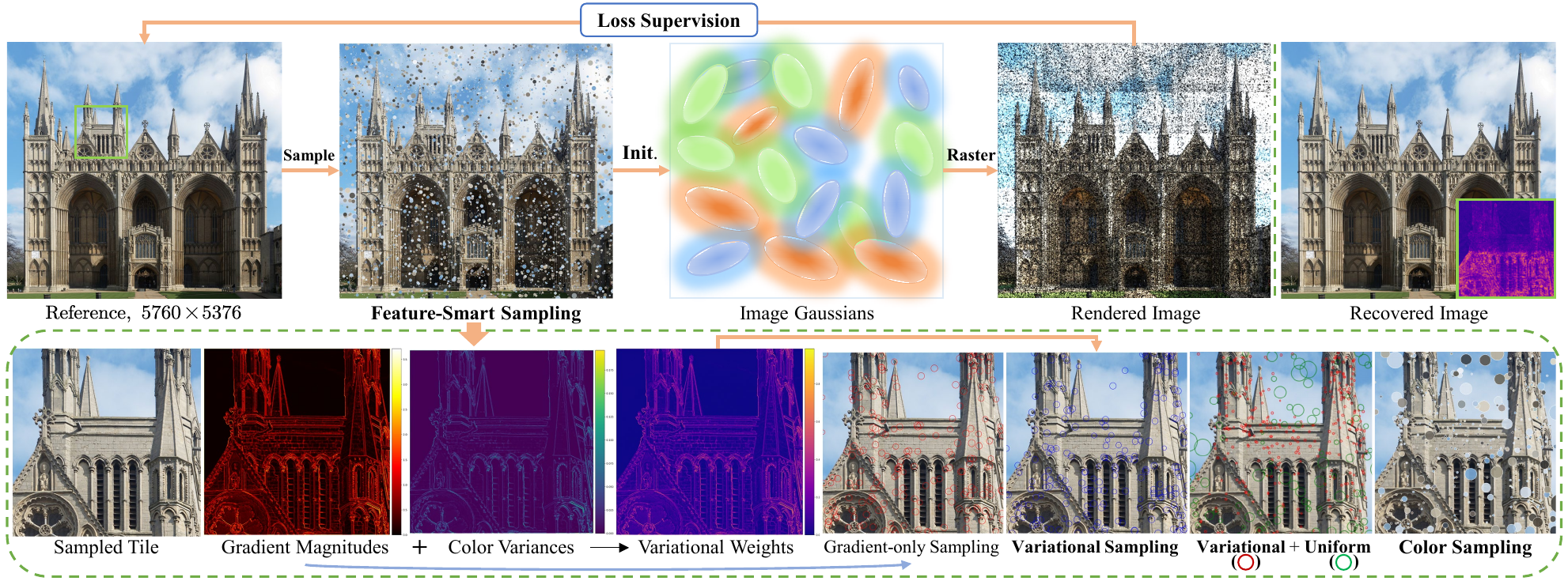}
    \caption{Pipeline of SmartSplat. Given an input image, SmartSplat begins by performing feature-smart sampling, where local image features, specifically gradient magnitudes and color variances, are analyzed to guide a variational sampling process. This process adaptively selects informative patches to initialize the positions, scales, and colors of a set of image-space Gaussians. These Gaussians are then passed through a differentiable rasterization pipeline, producing a rendered image. The system is supervised by a reconstruction loss computed between the rendered and original images, enabling gradient-based optimization of Gaussian parameters. Through this pipeline, SmartSplat learns a compact, content-aware Gaussian representation capable of reconstructing high-fidelity images under extreme compression constraints.}
    \label{fig:teaser_pipeline}
    \vspace{-10px}
\end{figure*}

\section{Adaptive Step-Size tiled Variational Sampling}

\subsection{Overview}
\textit{This section provides a supplementary explanation to the``Gradient-Color Guided Variational Sampling" subsection in the main text, offering a detailed exposition of the adaptive step-size tiled variational sampling strategy.}

To efficiently process ultra-high-resolution images while avoiding out-of-memory issues and ensuring uniform coverage during the initialization of Gaussian primitives, we propose an adaptive step-size tiled variational sampling strategy. This approach partitions the input image into multiple overlapping or adjacent tiles and performs variational sampling independently within each tile. By introducing adaptive strides for tile placement, the strategy ensures spatially uniform coverage across the entire image domain.

\subsection{Adaptive Tiling Strategy}
Given an input image of size $H \times W$, the numbers of tiles required along the height and width dimensions, denoted by $N_h$ and $N_w$, are computed as follows:
\begin{equation}
    \begin{aligned}
        N_h &= \max\left(1, \left\lceil \frac{H}{T} \right\rceil\right), \\
        N_w &= \max\left(1, \left\lceil \frac{W}{T} \right\rceil\right),
    \end{aligned}
\end{equation}
where $T$ represents the predefined tile size (set to 1024 in our experiments), and $\lceil \cdot \rceil$ denotes the ceiling function. To ensure uniform spatial distribution of tiles across the image, adaptive strides $s_h, s_w$ are subsequently computed as follows:
\begin{equation}
    \begin{aligned}
        s_h &=
            \begin{cases}
            0 & \text{if } N_h = 1 \\
            \frac{H - T}{N_h - 1} & \text{if } N_h > 1
            \end{cases}, \\
        s_w &=
        \begin{cases}
        0 & \text{if } N_w = 1 \\
        \frac{W - T}{N_w - 1} & \text{if } N_w > 1
        \end{cases}.
    \end{aligned}
\end{equation}

When only a single tile is required along a given dimension, the tile is positioned centrally within the image:
\begin{equation}
    \begin{aligned}
        p_h^{(0)} &= \max\left(0, \frac{H - T}{2}\right), \\
        p_w^{(0)} &= \max\left(0, \frac{W - T}{2}\right).
    \end{aligned}
\end{equation}
Otherwise, the position of the $i$-th tile along the height dimension is computed as:
\begin{equation}
    p_h^{(i)} = \min(i \cdot s_h, H - T),
\end{equation}
and the position of the $j$-th tile along the width dimension is given by:
\begin{equation}
    p_w^{(j)} = \min(j \cdot s_w, W - T),
\end{equation}
where $i = 0, \ldots, N_h - 1$ and $j = 0, \ldots, N_w - 1$.

\subsection{Tiled Variational Sampling}
Based on the aforementioned adaptive tiling strategy, the image patch corresponding to tile $(i, j)$ can be formally defined as:
\begin{equation}
    \mathbf{I}_{i,j} = \mathbf{I}\big[p_h^{(i)} : p_h^{(i)} + T_h^{(i,j)},\quad p_w^{(j)} : p_w^{(j)} + T_w^{(i,j)}\big],
\end{equation}
where the dimensions of the patch are given by:
\begin{equation}
    \begin{aligned}
        T_h^{(i,j)} &= \min(T,\, H - p_h^{(i)}), \\
        T_w^{(i,j)} &= \min(T,\, W - p_w^{(j)}).
    \end{aligned}
\end{equation}

Within each tile sub-image $\mathbf{I}_{i,j}$, the local gradient magnitude and color variance of its pixels are computed as follows:
\begin{equation}
    \begin{aligned}
m_{i,j}(\mathbf{x}) &= \frac{1}{C} \sum_{c=1}^C \left\| \nabla \mathbf{I}_{i,j,c}(\mathbf{x}) \right\|_2, \\
v_{i,j}(\mathbf{x}) &= \frac{1}{C} \sum_{c=1}^C \mathrm{Var}\bigl(\mathbf{I}_{i,j,c}(\mathcal{N}_{\mathbf{x}})\bigr),
\end{aligned}
\end{equation}
where $C$ denotes the number of channels, and $\mathcal{N}_{\mathbf{x}}$ represents the neighborhood of pixel $\mathbf{x}$. To eliminate scale discrepancies, the gradient magnitude and color variance are normalized within the tile:
\begin{equation}
    \begin{aligned}
       \tilde{m}_{i,j}(\mathbf{x}) &= \frac{m_{i,j}(\mathbf{x})}{\max_{\mathbf{x} \in \mathbf{I}_{i,j}} m_{i,j}(\mathbf{x}) + \epsilon}, \\
\tilde{v}_{i,j}(\mathbf{x}) &= \frac{v_{i,j}(\mathbf{x})}{\max_{\mathbf{x} \in \mathbf{I}_{i,j}} v_{i,j}(\mathbf{x}) + \epsilon}.
    \end{aligned}
\end{equation}
where $\epsilon$ is a small constant added for numerical stability. Then, the sampling weight is defined as a weighted combination of these normalized values:
\begin{equation}
    w_{i,j}(\mathbf{x}) = \lambda_m \cdot \tilde{m}_{i,j}(\mathbf{x}) + (1- \lambda_m) \cdot \tilde{v}_{i,j}(\mathbf{x}),
    \label{eq:vs_weight}
\end{equation}
where $\lambda_m$ denotes the weighting coefficient that balances the contributions of gradient magnitude and color variance. In our experiments, $\lambda_m$ is empirically set to 0.9 to achieve a favorable trade-off between structural detail and color distribution.

\subsection{Sampling Probability and Point Selection}
Based on the defined sampling weights, the probability of selecting a pixel $\mathbf{x}$ within tile $(i,j)$ is computed as:
\begin{equation}
    \mathbb{P}_{i,j}(\mathbf{x}) = \frac{w_{i,j}(\mathbf{x})}{\sum_{\mathbf{y} \in \mathbf{I}_{i,j}} w_{i,j}(\mathbf{y})}.
\end{equation}
Subsequently, $n_{i,j}$ pixels are sampled from each tile via multinomial sampling according to this probability distribution:
\begin{equation}
    \{\mathbf{x}_k^{(i,j)}\}_{k=1}^{n_{i,j}} \sim \mathrm{Multinomial}\left(n_{i,j}, \{\mathbb{P}_{i,j}(\mathbf{x})\}_{\mathbf{x} \in \mathbf{I}_{i,j}}\right).
    \label{eq:Multinomial}
\end{equation}
This sampling strategy promotes denser selection in regions exhibiting high gradient magnitudes or significant color variance, thereby enhancing the initialization quality of Gaussian primitives in perceptually salient areas.

\subsection{Sampling Allocation and Global Coordinate Conversion}
Assume the total number of variational sampling points, denoted by $N_g^{vs}$, is uniformly allocated to all tiles. For each tile located at $(i, j)$, the number of assigned samples $n_{i,j}$ in Eq. \ref{eq:Multinomial} is computed as:
\begin{equation}
    n_{i,j} = \left\lfloor \frac{N_g^{vs}}{N_h \times N_w} \right\rfloor + \mathbf{1}_{(i \times N_w + j) < (N_g^{vs} \bmod (N_h \times N_w))},
\end{equation}
where $\mathbf{1}_{\cdot}$ denotes the indicator function, which ensures an even distribution of the residual samples arising from modulo operation.

The sampled local coordinates $(\tilde{x}, \tilde{y})$ within each tile are subsequently converted to global image coordinates as follows:
\begin{equation}
    x_{\text{global}} = \tilde{x} + p_w^{(j)}, \quad y_{\text{global}} = \tilde{y} + p_h^{(i)},
\end{equation}
where $p_w^{(j)}$ and $p_h^{(i)}$ represent the horizontal and vertical offsets of tile $(i, j)$, respectively.

\subsection{Adaptive Scale Computation}
Evidently, points with higher sampling weights should be assigned smaller scales, while those with lower weights can be allocated larger scales. To ensure spatial smoothness, we adopt an exponential decay function to adaptively compute the scale. Assuming the initial scales along the $x$- and $y$-axes are equal, the scale is given by:
\begin{equation}
    s_{i,j}(\mathbf{x}) = s_{{base}} \cdot \exp(-\frac{1}{2}w_{i,j}(\mathbf{x})).
    \label{eq:scale_exp}
\end{equation}

Assuming that each initialized Gaussian exhibits isotropic scaling (i.e., equal lengths of the major and minor axes), and that the image domain of size $H \times W$ is uniformly partitioned by $N_g$ non-overlapping circles, the maximum radius $R_{max}$ of each circle can be derived based on the principle of equal-area coverage:
\begin{equation}
R_{max} = \sqrt{\frac{H W}{\pi N_g}}.
\end{equation}
To ensure maximal spatial coverage of the image while accounting for the effective influence radius of each Gaussian during rasterization, the base scale $s_{base}$ in Eq. \ref{eq:scale_exp} is further defined as one-third of $R_{max}$:
\begin{equation}
s_{base} = \frac{1}{3} R_{g \rightarrow p} = \frac{1}{3} R_{max} = \frac{1}{3} \sqrt{\frac{H W}{\pi N_g}}.
\end{equation}
This scale initialization strategy enables adaptive representation of images at arbitrary resolutions, without requiring any additional hyperparameters or heuristic clamping.

\begin{figure*}[ht]
    \centering
    \includegraphics[width=0.96\textwidth]{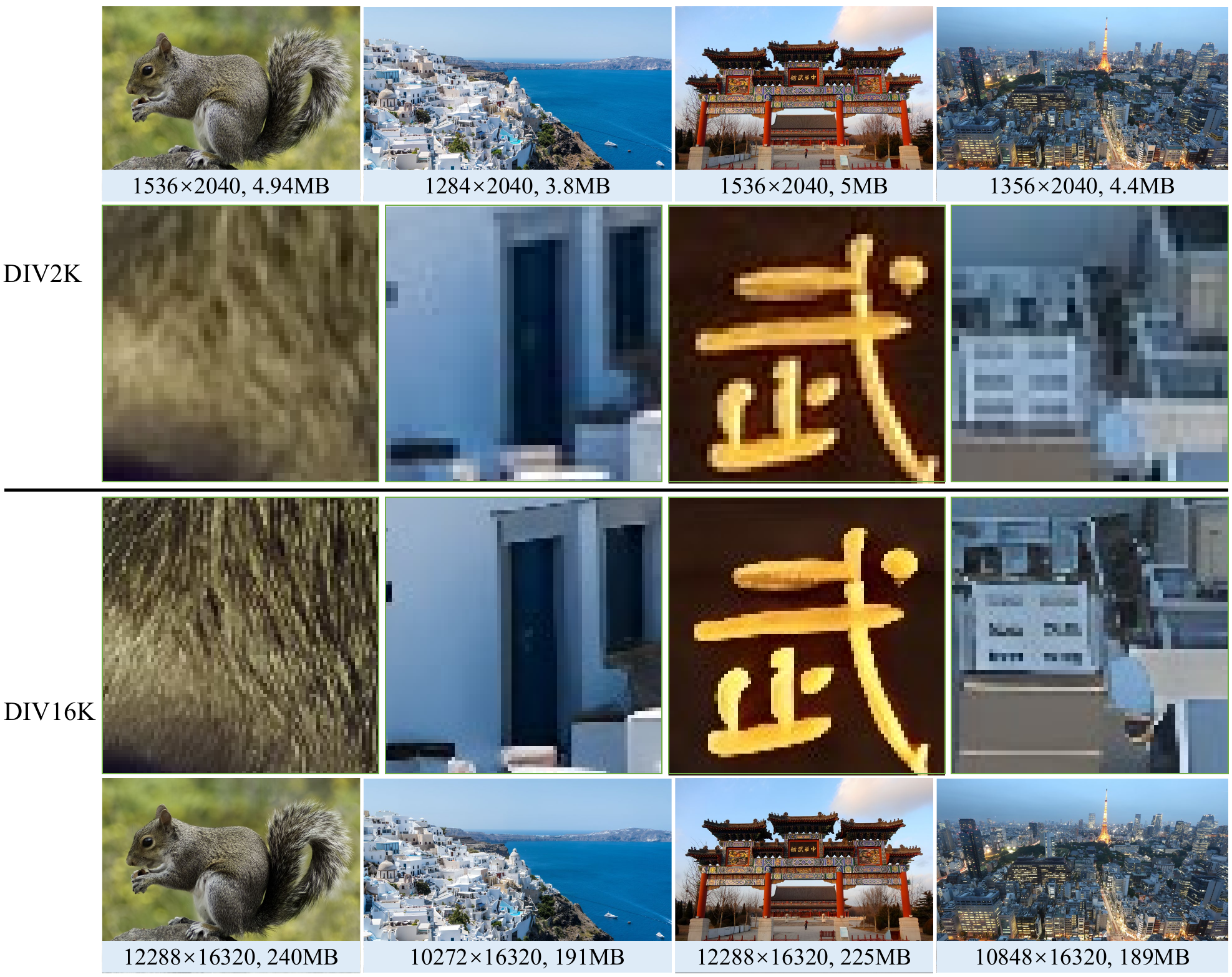}
    \caption{Comparison between DIV16K and DIV2K Samples.  By applying an 8$\times$ upsampling using the Aiarty Image Enhancer to DIV2K images, our created DIV16K dataset exhibits notably clearer details upon local zoom-in. Nevertheless, the ultra-high-resolution images impose substantial storage demands. This dataset can serve as a valuable benchmark for future research in AI-based ultra-high-resolution image generation and processing, offering significant practical and scientific relevance.}
    \label{fig:teaser_div16k}
    \vspace{-10px}
\end{figure*}

\section{DIV16K Dataset}
\textit{This section provides a detailed description of the constructed DIV16K dataset and serves as a supplementary explanation to the ``Dataset" subsection within the ``Experimental Setup" section of the main paper.}

As illustrated in Fig. \ref{fig:teaser_div16k}, to address the challenges posed by the storage and transmission of AI-generated ultra-high-resolution images, this study constructs the DIV16K dataset based on the DIV2K \cite{div2k2017} dataset by applying an 8$\times$ upsampling using the Aiarty Image Enhancer, resulting in 800 images at 16K resolution. In conventional formats such as PNG or JPEG, the storage size of these images typically ranges from 100 MB to 300 MB per image, imposing significant burdens on storage and network transmission. By leveraging the SmartSplat method, it is possible to substantially reduce storage requirements while maintaining high-fidelity image representations. Detailed visualizations of our method’s performance in image compression and reconstruction are available on the project website (https://smartsplat.github.io/SmartSplat-Website/).

\section{Experimental Details}
\subsection{Implementation.}
To mitigate the memory overhead of UHR images during initialization, all sampling procedures were implemented in a tile-based manner. For uniform sampling, we designed a CUDA-based query-to-reference KNN pipeline that enables efficient exclusion sampling and scale estimation over large Gaussian points. On 16K-resolution images, the initialization stage can be completed within approximately $2 \sim 5$ seconds. In variational sampling, the weight $\lambda_m$ was set to 0.9. For uniform sampling, the parameter $K$ was set to 3. During training, the proportion of variational sampling $\lambda_g$ was 0.7, and the loss weight $\lambda_l$ was set to 0.9. All Gaussian parameters (means, scales, colors and rotation angles) were jointly optimized using the Adam optimizer over 50,000 steps, with learning rates of $1e-4$, $5e-3$, $5e-2$, and $1e-3$, respectively. Due to the lack of batch parallelism support in GS rasterization, all experiments and evaluations were conducted on a single GPU within an A800 (80GB) cluster.

\subsection{Evaluation Metrics.}
Peak Signal-to-Noise Ratio (PSNR) is employed to quantify pixel-level distortion between the reconstructed and ground truth images. To more comprehensively assess perceptual quality and structural fidelity, Multi-Scale Structural Similarity Index (MS-SSIM) \cite{Wang2003msssim} is adopted as a structural error metric, particularly suitable for 8K-resolution images. However, due to the risk of OOM errors when computing MS-SSIM on 16K ultra-high-resolution images, we instead utilize an efficient implementation of SSIM proposed by \cite{Mallick2024fusedssim}, based on the original SSIM formulation \cite{Zhou2004ssim}, to ensure stable evaluation.

\begin{figure*}[ht]
    \centering
    \includegraphics[width=0.96\textwidth]{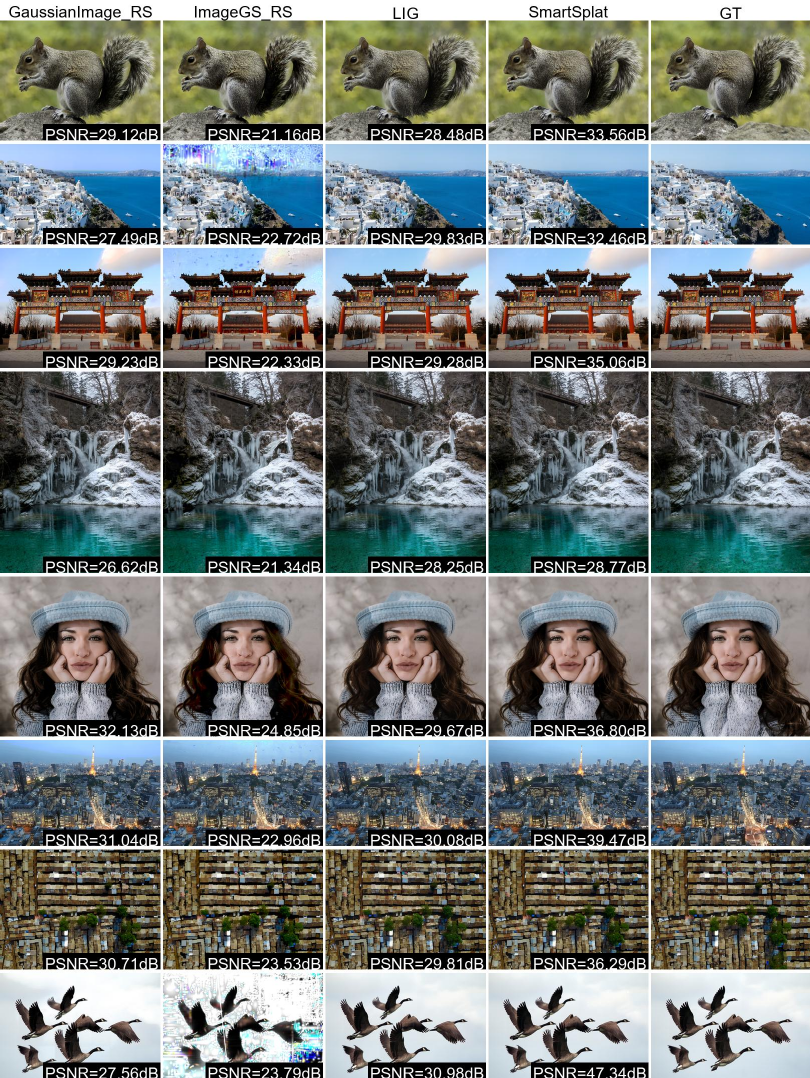}
    \caption{Qualitative results on DIV16K. (CR = 50)}
    \label{fig:div16k_cr50}
\end{figure*}

\begin{figure*}[ht]
    \centering
    \includegraphics[width=0.96\textwidth]{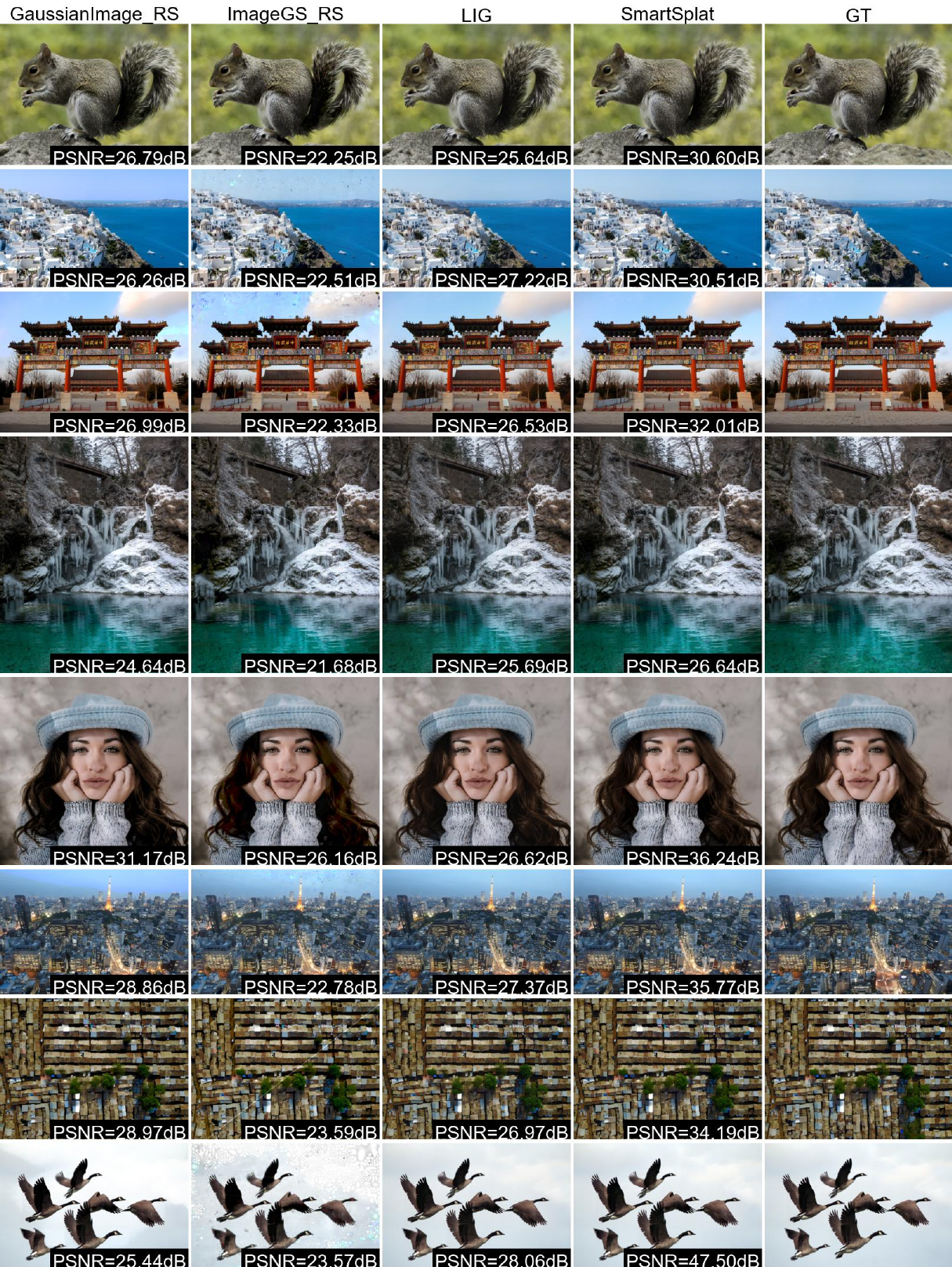}
    \caption{Qualitative results on DIV16K. (CR = 100)}
    \label{fig:div16k_cr100}
\end{figure*}

\begin{figure*}[ht]
    \centering
    \includegraphics[width=0.96\textwidth]{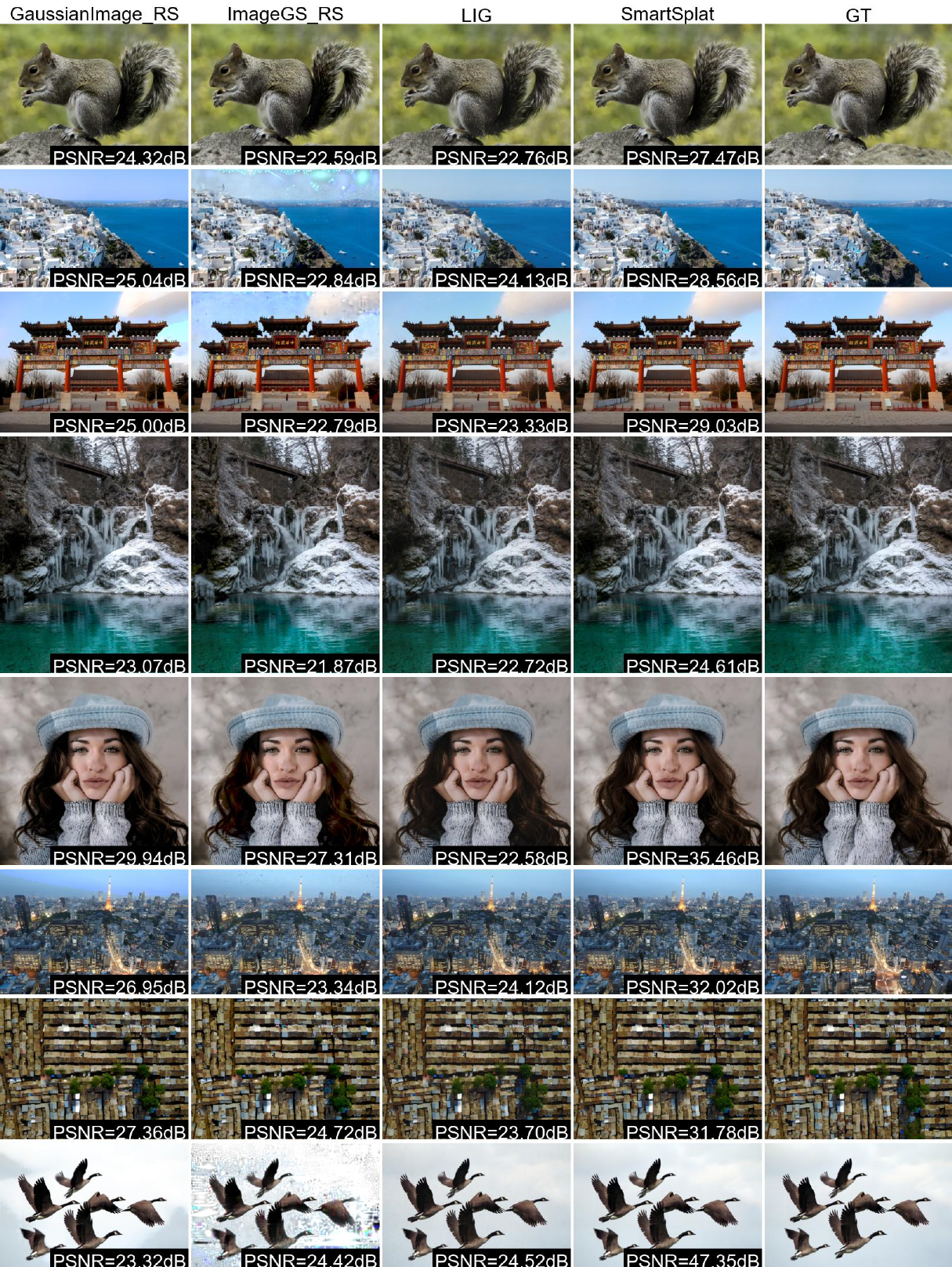}
    \caption{Qualitative results on DIV16K. (CR = 200)}
    \label{fig:div16k_cr200}
\end{figure*}

\begin{figure*}[ht]
    \centering
    \includegraphics[width=0.8\textwidth]{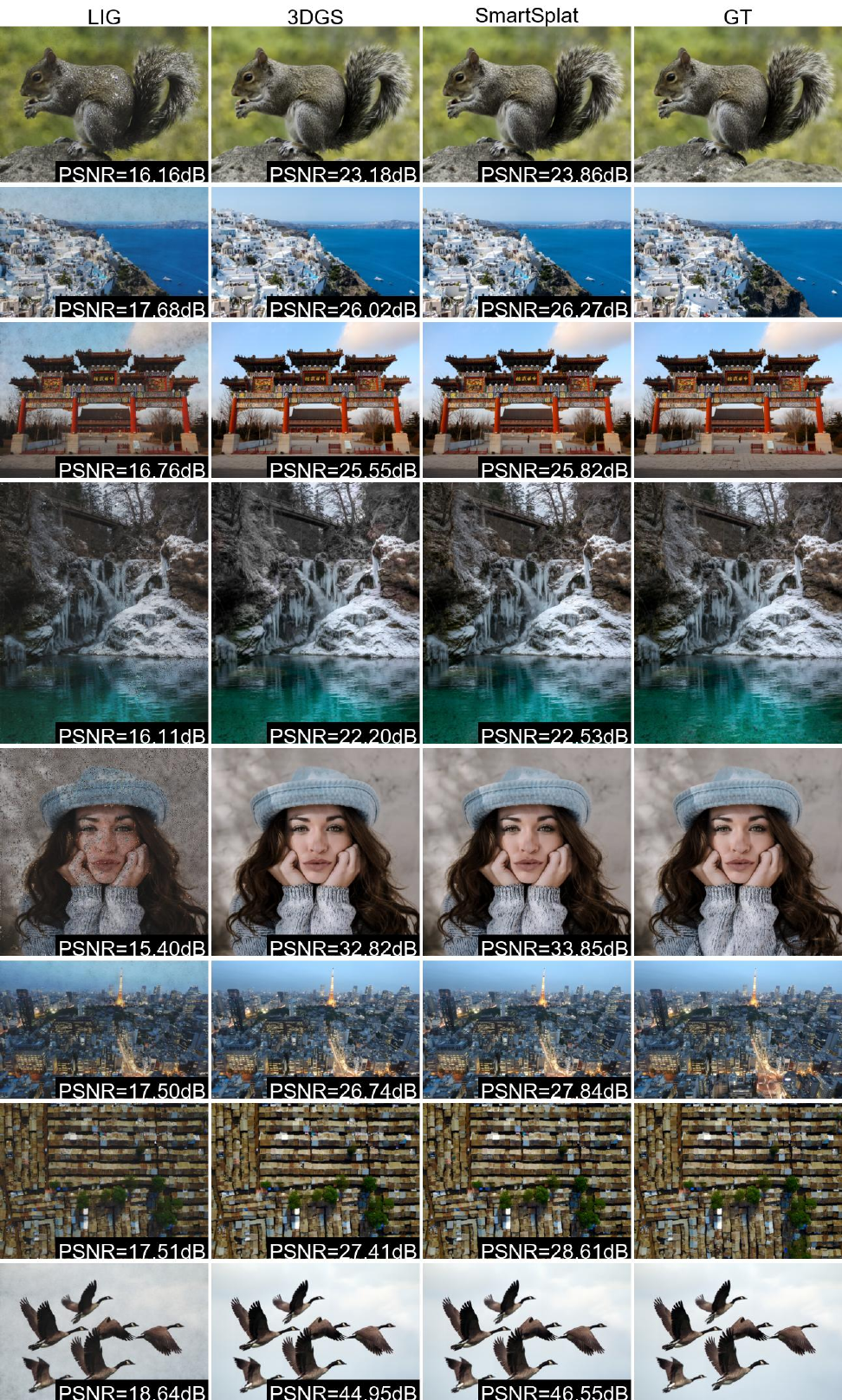}
    \caption{Qualitative results on DIV16K. (CR = 500)}
    \label{fig:div16k_cr500}
\end{figure*}

\begin{figure*}[ht]
    \centering
    \includegraphics[width=0.96\textwidth]{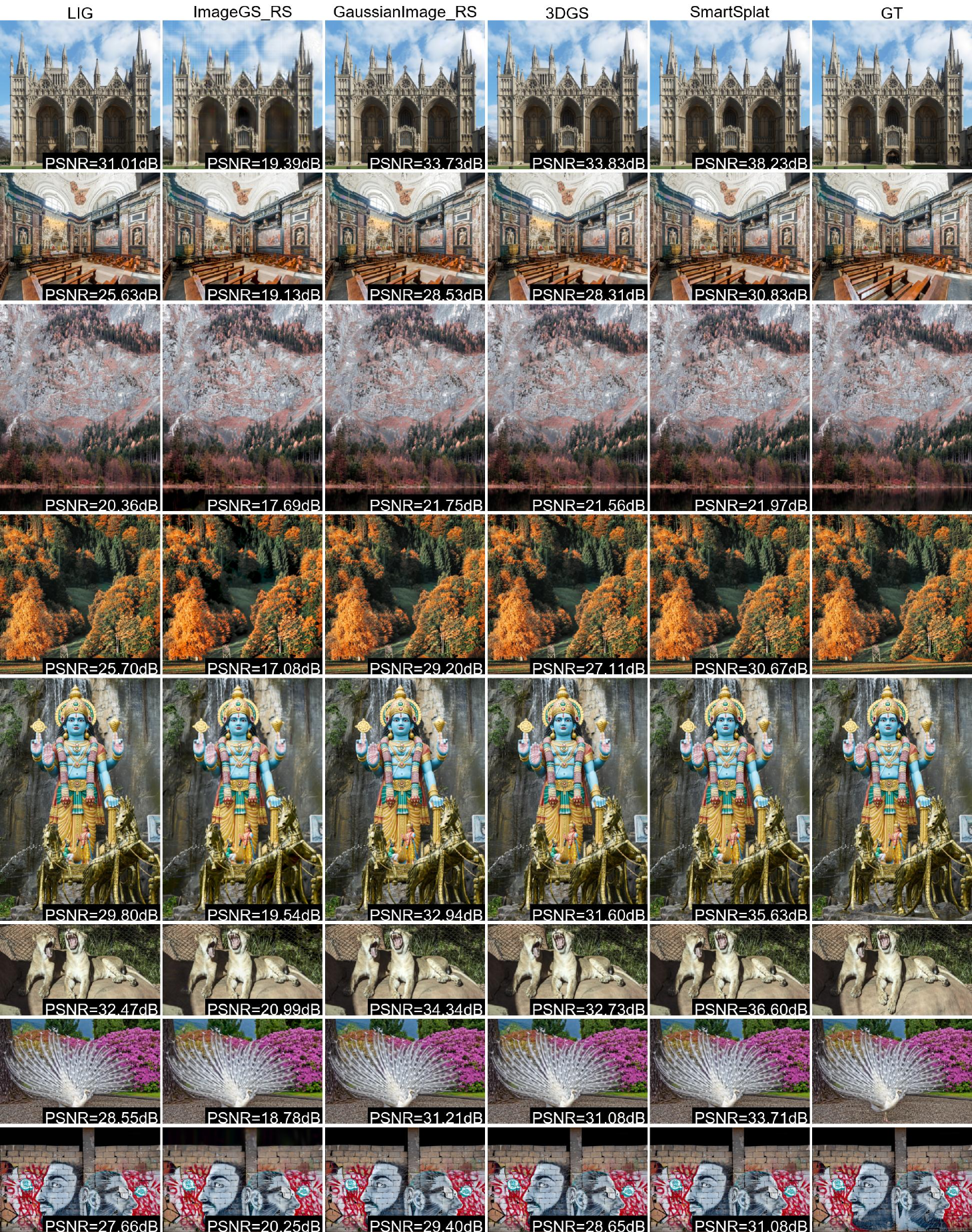}
    \caption{Qualitative results on DIV8K. (CR = 20)}
    \label{fig:div16k_cr20}
\end{figure*}

\begin{figure*}[ht]
    \centering
    \includegraphics[width=0.96\textwidth]{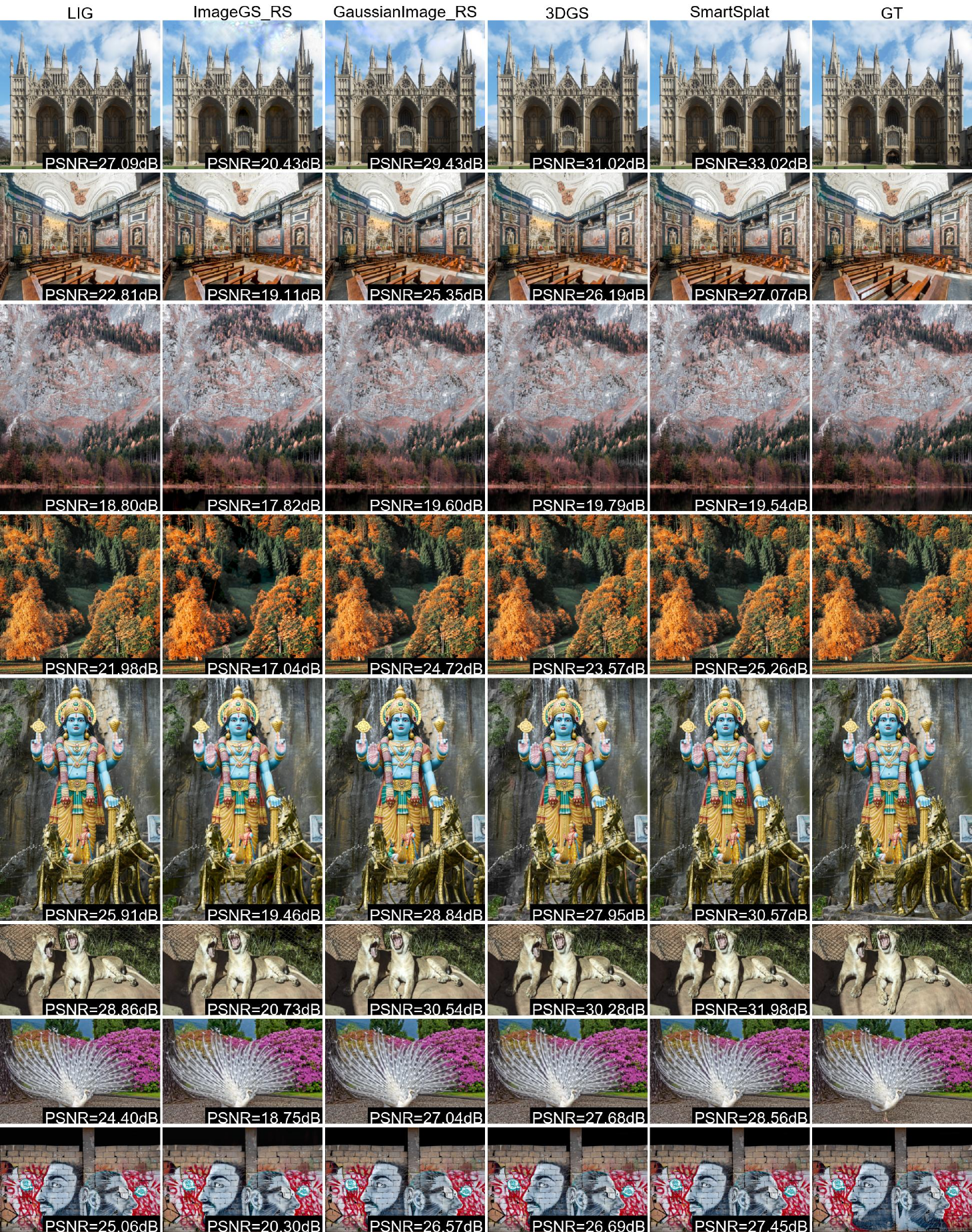}
    \caption{Qualitative results on DIV8K. (CR = 50)}
    \label{fig:div16k_cr50}
\end{figure*}

\begin{figure*}[ht]
    \centering
    \includegraphics[width=0.96\textwidth]{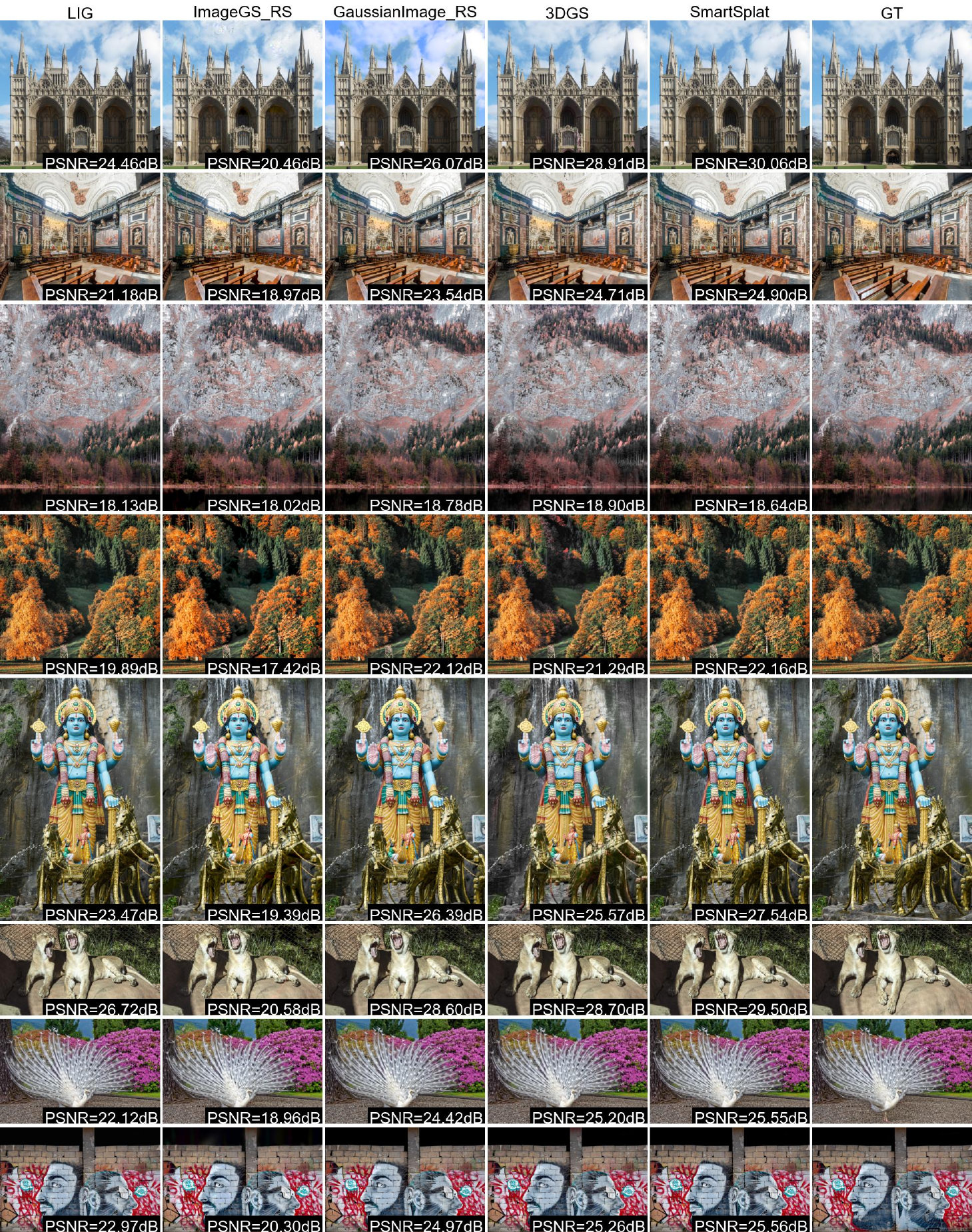}
    \caption{Qualitative results on DIV8K. (CR = 100)}
    \label{fig:div16k_cr100}
\end{figure*}

\begin{figure*}[ht]
    \centering
    \includegraphics[width=0.96\textwidth]{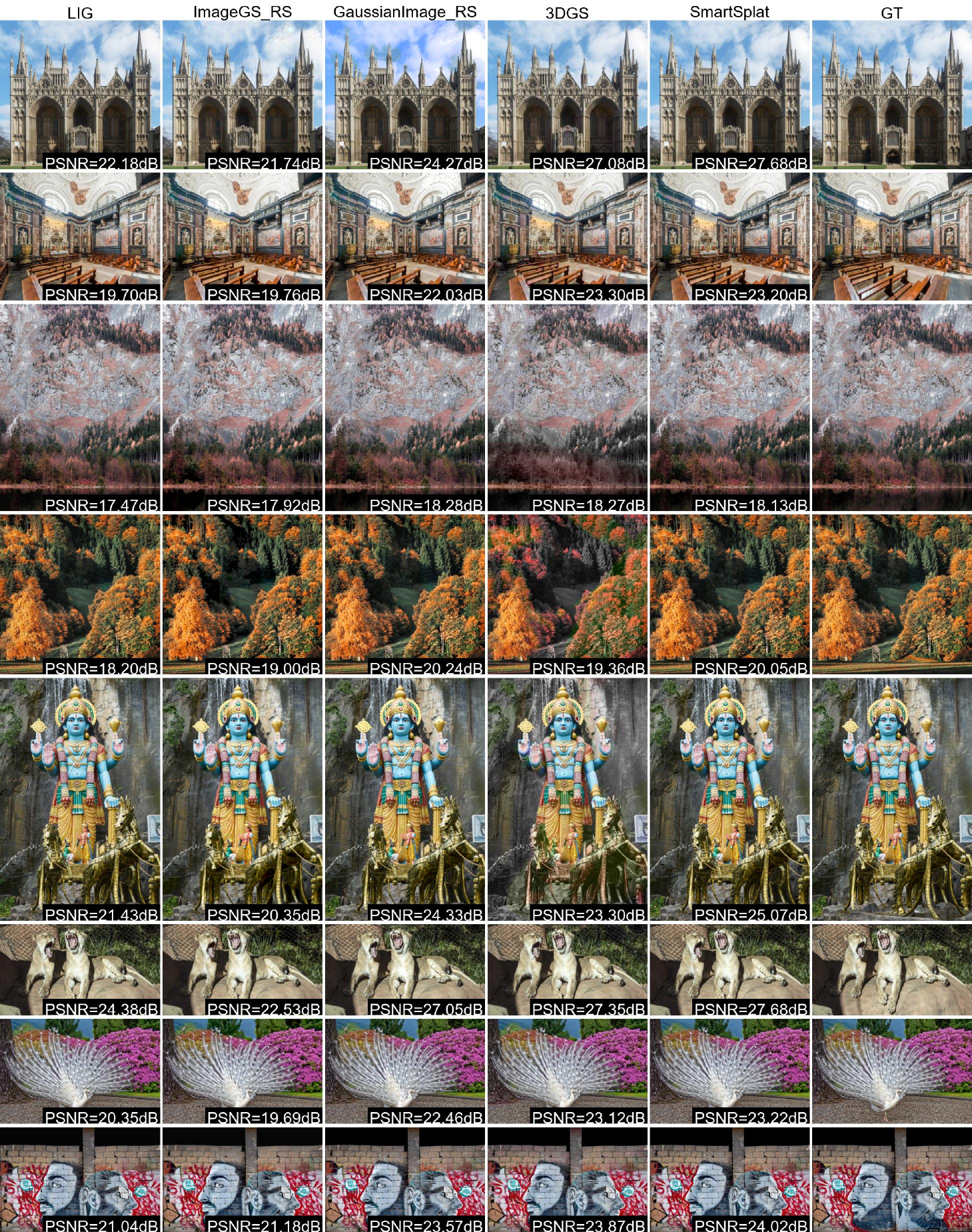}
    \caption{Qualitative results on DIV8K. (CR = 200)}
    \label{fig:div16k_cr200}
\end{figure*}

\end{document}